\documentclass{ieeeaccess}
\usepackage{cite}
\usepackage{amsmath,amssymb,amsfonts}
\usepackage{algorithmic}
\usepackage{graphicx}
\usepackage{caption}%
\usepackage{url}
\usepackage{textcomp}
\usepackage{algorithm}
\usepackage{algorithmic} 
\def\BibTeX{{\rm B\kern-.05em{\sc i\kern-.025em b}\kern-.08em
    T\kern-.1667em\lower.7ex\hbox{E}\kern-.125emX}}
\begin{document}
\history{Date of publication xxxx 00, 0000, date of current version xxxx 00, 0000.}
\doi{10.1109/ACCESS.2017.DOI}

\title{An Extensive and Methodical Review of Smart Grids for Sustainable Energy Management-Addressing Challenges with AI, Renewable Energy Integration and Leading-edge Technologies}

\author{\uppercase{Parag Biswas}\authorrefmark{1}, 
\uppercase{Abdur Rashid}\authorrefmark{1}, 
\uppercase{abdullah al masum}\authorrefmark{1}, 
\uppercase{MD Abdullah Al Nasim}\authorrefmark{2},
\uppercase{A.S.M Anas Ferdous}\authorrefmark{4},
\uppercase{Kishor Datta Gupta}\authorrefmark{3},
\uppercase{Angona Biswas}\authorrefmark{2}
}

\address[1]{MSEM Department, Westcliff University, California, United States (e-mail: text2parag@gmail.com; rabdurrashid091@gmail.com; a.masum.642@westcliff.edu)}
\address[2]{Research and Development Department, Pioneer Alpha, Dhaka, Bangladesh (e-mail: nasim.abdullah@ieee.org; ngonabiswas28@gmail.com)}
\address[3]{Department of Computer and Information Science, Clark Atlanta University, Georgia, USA (e-mail: kgupta@cau.edu)}
\address[4]{Department of Biomedical Engineering, Bangladesh University of Engineering and Technology, Dhaka, Bangladesh  (e-mail: anasferdous001@gmail.com)}

\corresp{Corresponding author: MD Abdullah Al Nasim (e-mail: nasim.abdullah@ieee.org).}

\begin{abstract}
Energy management decreases energy expenditures and consumption while simultaneously increasing energy efficiency, reducing carbon emissions, and enhancing operational performance. Smart grids are a type of sophisticated energy infrastructure that increase the generation and distribution of electricity's sustainability, dependability, and efficiency by utilizing digital communication technologies. They combine a number of cutting-edge techniques and technology to improve energy resource management. A large amount of research study on the topic of smart grids for energy management has been completed in the last several years. The authors of the present study want to cover a number of topics, including smart grid benefits and components, technical developments, integrating renewable energy sources, using artificial intelligence and data analytics, cybersecurity, and privacy. Smart Grids for Energy Management are an innovative field of study aiming at tackling various difficulties and magnifying the efficiency, dependability, and sustainability of energy systems, including: 1) Renewable sources of power like solar and wind are intermittent and unpredictable 2) Defending smart grid system from various cyber-attacks 3) Incorporating an increasing number of electric vehicles into the system of power grid without overwhelming it. Additionally, it is proposed to use AI and data analytics for better performance on the grid, reliability, and energy management. It also looks into how AI and data analytics can be used to optimize grid performance, enhance reliability, and improve energy management. The authors will explore these significant challenges and ongoing research. Lastly, significant issues in this field are noted, and recommendations for further work are provided.

\end{abstract}


\titlepgskip=-15pt

\maketitle

\section{Introduction}
\label{sec:introduction}

Electricity is a crucial energy source for addressing the needs of both individuals in developing countries and fostering economic growth. Reliable and efficient power transmission is essential for a nation's economic health \cite{gegner2017phasor}. Since CO2 emissions significantly contribute to global warming, targeting the major sources of these emissions, such as transportation and power generation, is key to addressing climate change \cite{nair2018assessment}. A viable route to a sustainable future is the switch to electric vehicles (known as EVs) and renewable energy sources (in short RESs). Moreover, integrating energy storage systems (ESSs) or battery energy storage systems (known as BESSs) into the current grid and replacing conventional fuel-based transportation with electric alternatives such as plug-in hybrid electric vehicles (known as PHEV) as well as plug-in electric vehicles (in short PEVs) are workable strategies to reduce the sharp rise in greenhouse gas (GHG) emissions \cite{khan2022energy}. Unlike conventional energy sources, renewable energies are characterized by their variability and intermittency.

The variability of RES can be managed by integrating multiple RES with energy storage systems (ESS) and backup sources \cite{rana2023applications}. However, this variability can significantly impact the system's voltage stability, disrupt traditional on-load tap changer control mechanisms, and adversely affect the overall performance of the power grid. An energy management system (EMS) is essential for optimizing the potential of new resources and types of loads on the electricity network, reducing their negative impacts, ensuring continuous load supply under all conditions, and enhancing network stability. According to the International Electrotechnical Commission's IEC 61970 standard \cite{rathor2020energy}, an EMS is defined as "a computer system comprising a software platform that offers essential support services and a set of applications necessary for the efficient operation of generation and transmission systems to ensure energy supply security at minimal cost." Energy management in the framework of a smart grid (in short SG) guarantees supply and demand balance while adhering to all system restrictions to achieve economical, dependable, and secure electrical system operation \cite{meliani2021energy}.

\subsection{A Long View of Energy Management: From Fire to Smart Grids}
While the concept of modern energy management is recent, humanity's quest for efficient power use stretches back millennia.  Early humans conserved fire for warmth, and later harnessed wind and water for rudimentary machines, demonstrating a constant drive to optimize resource utilization \cite{black2022have}.  The development of the steam engine in the 18th century marked a major turning point \cite{dickinson2022short}. James Watt's design prioritized efficiency, enabling a significant reduction in fuel consumption for the same power output \cite{dickinson2022short}.

The true birth of modern energy management, however, is attributed to the oil crisis of the 1970s. This event spurred a shift from simply using less energy to actively managing it \cite{wellum2023energizing}. The 1980s saw the rise of energy audits, monitoring systems, and efficiency upgrades, reflecting a focus on proactive management \cite{turnbull2022no}.  This period also witnessed the introduction of micro-combined heat and power (CHP) technology in 1981, offering efficient on-site generation, and the launch of the Energy Efficiency Year campaign in the UK (1986).

The focus then transitioned towards strategic energy procurement, before environmental concerns in the 2000s drove a shift towards carbon reduction and renewable energy sources \cite{o2021corporate}. In order to achieve sustainability and efficiency, data analytics and intelligent technologies are now reshaping the field of energy management \cite{o2019smart}, \cite{bibri2021novel}. Table \ref{tab:energy_management_system} summarizes existing literature on various energy management strategies, outlining their contributions in types, architecture, components, stakeholders, solution approaches, and programs like demand response (known as DR) and demand-side management (known as DSM). The table also identifies systems such as building energy management systems (known as BEMS), energy management systems (known as EMS), smart urban energy system (SUES), energy storage systems (known as ESS), home energy management systems (known as HEMS), and plug-in electric vehicles (known as PEV).

\begin{table*}[htbp]
\centering
\caption{Summary of Existing Literature on Energy Management Systems}
\resizebox{\textwidth}{!}{
\begin{tabular}{|c|ccc|cccc|cccc|}
\hline

\multicolumn{1}{|r}{} & \multicolumn{3}{|c|}{Categories} & \multicolumn{4}{c|}{System Design} & \multicolumn{4}{c|}{Focus Areas} \\ \cline{2-12}

\hline
Sr. No. & References & Types & Architecture & Solar & Wind & ESS & PEV & Stakeholders & DR/DSM & Approaches for solution & Published Year \\ \hline
1 & Ref. \cite{zheng2024systematic} & SG and SUES & &  &  & & & Yes & Yes & Yes & 2024 \\

2 & Ref. \cite{olatunde2024impact} & SG & Yes &  & &  & & Yes & Yes & 
Yes & 2024 \\

3 & Ref. \cite{bakare2023comprehensive} & DSM & & & & & & & Yes &  Yes & 2023 \\

4 & Ref. \cite{poompavai2019control} & EMS & & Yes & Yes & & & & & & 2019 \\
5 & Ref. \cite{azuatalam2019energy} & EMS & & Yes & & Yes & & & & 
Yes & 2019 \\
6 & Ref. \cite{yousefi2019comparison} & EMS & & & & & & & &  Yes & 2019 \\
7 & Ref. \cite{zou2019survey} & EMS & & & Yes & & & & & Yes & 2019 \\
8 & Ref. \cite{cheng2018centralize} & EMS & Yes & Yes & Yes & Yes & & & & & 2018 \\
9 & Ref. \cite{shareef2018review} & HEMS & Yes & & & & & & Yes & Yes & 2018 \\
10 & Ref. \cite{hannan2018review} & BEMS & & & & & & Yes & & & 2018 \\
11 & Ref. \cite{zia2018microgrids} & EMS & & & & & & Yes & & Yes & 2018 \\
12 & Ref. \cite{vivas2018review} & EMS & & Yes & Yes & Yes & & & & & 2018 \\
13 & Ref. \cite{zakaria2020uncertainty} & BEMS & & & & & & & & & 2019 \\
14 & Ref. \cite{khan2019optimal} & EMS & Yes & & & & & & & Yes & 2018 \\
15 & Ref. \cite{salimi2019critical} & BEMS & & & & Yes & & & & & 2018 \\
\hline
\end{tabular}
}
\label{tab:energy_management_system}
\end{table*}

\subsection{What exactly does energy management entail?}

Energy management has emerged as a systematic approach to overseeing the entire energy lifecycle, encompassing production, distribution, and consumption \cite{vlachokostas2020closing}.  Its core objective is to achieve  energy efficiency, resulting in reduced costs and environmental impact.  This is accomplished by combining methods and tools aimed at increased productivity, including renewable energy sources, and employing demand control techniques.

A successful energy management program relies on a robust framework encompassing several key components.  Energy auditing serves as the foundation, providing a comprehensive assessment of current energy use and pinpointing areas for improvement.  Energy monitoring and data analysis leverage specialized tools and systems to track real-time consumption and historical trends, revealing further optimization opportunities.  Energy conservation translates insights into action through behavioral changes, process adjustments, and equipment upgrades to reduce overall energy usage \cite{zhou2016understanding}.  Optimization of energy systems focuses on enhancing the performance of existing systems, such as HVAC, lighting, and industrial processes, to operate more efficiently \cite{simpeh2022improving}, \cite{gonzalez2018multi}.

In addition, the incorporation of energy from renewable sources such as wind and solar power is essential for decreasing dependency on fossil fuels and lowering greenhouse gas emissions. Demand response strategies proactively manage energy demand to align with supply, particularly during peak periods, by encouraging users to adjust their consumption patterns.  Energy storage solutions utilize technologies like batteries to capture excess energy production for use when demand is high or renewable energy sources are limited.  Finally, sustainability planning establishes a long-term vision to ensure responsible energy use, meeting regulatory requirements and aligning with corporate goals for environmental stewardship.

\subsection{The Smart Grid Concept in Management of Energy}

Although the idea of a sharper electrical grid has been around for decades, widespread adoption has not happened yet. The efficiency, dependability, and capacity to integrate renewable energy sources into the conventional electrical grid—a complicated web of generating plants, lines for transmission, and distribution networks—have their limits. Using electronic devices and two-way communication to create a more sophisticated and responsive energy infrastructure, the smart grid hypothesis emerges as a game-changing solution.  

The smart grid concept dates back to the late nineteenth century, when Nikola Tesla proposed a more adaptable alternating current (AC) power system. The twentieth century paved the way for the smart grid revolution \cite{maloberti2022short}. Advances in communication technology, like as SCADA systems designed for remote monitoring and control, provided the opportunity for real-time grid management \cite{katyara2019monitoring}.  Sensor technology advancements drove this idea by allowing for the capture of critical data from multiple places on the grid.  Furthermore, an increasing emphasis on energy conservation and environmental concerns during this time period fueled the demand for a more intelligent and adaptive grid infrastructure, paving the way for the creation of the smart grid idea. The late twentieth century saw the beginning of the smart grid era. Deregulation of the energy sector, notably in the United States, has created opportunities for innovation and investment in new technologies. This transition prompted an increase in research and development activities, with a particular emphasis on integrating digital communication and automation capabilities into the existing power grid infrastructure. The 21st century saw a new era of actual action for the smart grid. Governments, such as the US Department of Energy, have provided major financing for pilot programs, hastening smart grid development.  Small-scale adoption of these technologies was initiated by utilities and technology companies, who initially concentrated on solutions aimed at consumers, such as smart meters as well as automated distribution networks inside the grid. The smart grid is still developing today. Now that these technologies are being developed and implemented, incorporating renewable energy sources like solar and wind power is being prioritized once again. A sustainable energy future depends on this integration \cite{husin2021critical}, \cite{al2022sustainable}. Additionally, improving grid resilience to cyberattacks and natural catastrophes \cite{hossain2021metrics} is a significant concern.  Furthermore, providing consumers with real-time energy use statistics and dynamic pricing alternatives is a primary priority \cite{xu2021resilience}. Standardization across multiple systems, cost-cutting measures, and handling increasing regulatory challenges are all essential areas that need to be developed further with a view to realize full potential of smart grid.

\subsubsection{Mandatory Elements of Smart Grid for Energy Management}
A smart grid is a complex energy management system that makes use of contemporary technologies and communication infrastructure to optimize the whole energy lifecycle. Here are some required aspects of a smart grid:

\begin{itemize}
\item Advanced Metering Infrastructure (in short AMI): This replaces conventional meters with digital or smart meters that accumulate real-time data on energy consumption \cite{gold2020leveraging}, enabling: Detailed Usage Monitoring, Billing Accuracy and Demand Response Programs \cite{marimuthu2018development}. For monitoring, consumers can track their energy use by time of day, identifying areas for potential savings. More accurate billing system based on actual usage patterns. It utilities can incentivize consumers to shift energy use to off-peak periods.

\item Communication Network: Data transmission between smart grid components requires a secure, dependable two-way communication network. This network supports real-time monitoring, automated control, and data exchange \cite{khan2019secure}.  Real-time Monitoring utilities can assess grid health, identify possible problems, and optimize power flows \cite{dileep2020survey}. Automated Control is configured to allow remote control of grid components like as transformers and switches for increased efficiency \cite{dileep2020survey}. Data exchange is another crucial goal that allows customers, utilities, and energy providers to communicate with one another.

\item Smart Substations and Distribution Systems: Traditional substations and distribution systems are equipped with intelligent devices that collect data, automate processes, and enhance grid resilience \cite{chaves2022development}. These developments provide self-healing capabilities, which allow the grid to autonomously detect and isolate issues, reducing service interruptions. They also help with power quality control by managing voltage and frequency to ensure optimal equipment operation \cite{ge2022smart}. The ability to seamlessly integrate distributed energy resources (in short DERs) like solar and wind electricity into the grid is perhaps its most important function.

\item Smart Appliances and Consumer Engagement: Through the use of energy management programs and smart appliances, consumers contribute significantly to a smart grid.  Among these characteristics is automatic demand response, which enables appliances to adjust their energy use in response to dynamic pricing and grid conditions. Once more, it offers consumer education, enabling users to develop awareness of their energy usage and make informed choices on energy conservation. It also increases demand flexibility; consumers may engage in demand-response programs, which reduce peak demand on the grid.

\item Cybersecurity Measures: With an increasing reliance on digital communication, strong cybersecurity measures are required to safeguard the smart grid from cyberattacks that might interrupt energy supply. This comprises data encryption, access control, and cyber security protocols. Data Encryption uses encryption techniques to protect sensitive information. Access Control imposes strong access restrictions to prevent unwanted changes and cybersecurity procedures establishs comprehensive cybersecurity procedures and incident response strategies.

\item Regulatory Framework and Standards: A clearly identified regulatory framework and technology standardization are critical for the proper functioning and interoperability of diverse smart grid components. This includes standard communication protocols, data privacy regulations, and incentives for investment in smart grid technologies.

\end{itemize}

This research study on the topic of Smart Grids for Energy Management contributes in various ways:
\begin{itemize}
\item We comprehensively examine smart grid benefits, components, technological advancements, and their significance in integrating renewable energy sources.
\item The study focuses on critical challenges for smart grids, such as the irregular nature of renewable energy, cybersecurity concerns, as well as grid integration of electric vehicles.
\item We try to get to the bottom of the potential of AI and data analytics in addressing these issues and enhancing smart grid operations.
\item By assessing current research and highlighting key concerns, the study provides valuable recommendations for future research areas in this rapidly growing field.
\end{itemize}

\subsection{Integration of AI and Machine Learning}

Artificial Intelligence has emerged as one of the key technologies in the smart grid domain. AI-powered energy management systems will increase grid efficiency through optimized flow of energy, minimizing losses, and reducing wastage \cite{marques2024artificial}. The AI technologies allow for a more precise forecast of energy demand and generation, thus enabling better planning and operation of the grid. This cuts down on fossil fuel use and, consequently, reduces greenhouse gas emissions, hence contributing to environmental sustainability. Machine learning algorithms can be used to forecast renewable energy generation, analyze power flow, and optimize storage and distribution systems \cite{marques2024artificial}.

\section{Literature Review}

\subsection{Challenges in Modern Power Grids}

With its extensive network of plants of power, lines for transmission, and distribution systems, the present power grid is finding it difficult to meet our expanding energy needs.  According to a research by the International Energy Agency, there will be a 30\% increase in global energy consumption by 2040 \cite{saleem2023integrating}. This challenge is driving the development of smarter grids.  In Pakistan, for instance, monthly electricity demand is expected to jump from 120,000 GWh in 2019-2020 to 200,000 GWh by 2024-2025, highlighting the need for a more efficient system [2]. To address these challenges, researchers are actively exploring ways to improve energy management systems and integrate them with smart grid technologies for a more sustainable and reliable future. 

\subsection{Energy Management Systems (EMS), Demand Response Programs (DRP) and The Role of IoT and Cloud Computing in Smart Grids }
The study \cite{hussain2021multi} report offers the most recent information on various energy administration platforms at the network, aggregator, and household levels. Each system's benefits and drawbacks are examined and contrasted, taking into account its primary components—objective functions, limitations, algorithms for optimization, communication protocols, and the effect of EVs. Home EMS (HEMS) refers to energy management done at the house level, and Grid EMS (GEMS) refers to energy management done on the grid side. By utilizing the adaptability of the controlled loads at home, the HEMS lowers the cost of power. The GEMS is primarily focused on minimizing power losses, operating costs, and voltage control. Program for Demand Response (DRP) In order to balance a customer's power consumption and the utility's power supply, DRP modifies the consumer's demand for electricity on the energy grid. DRP is a crucial part of EMS that reduces peak demand, enhances load profiles, and permits greater utilization of system resources. DRP improves power system performance, minimizes voltage variation, optimizes income, and lowers operating expenses in addition to ensuring a dependable power system.


\begin{figure}[h]
\centering
\includegraphics[height= 5.5cm]{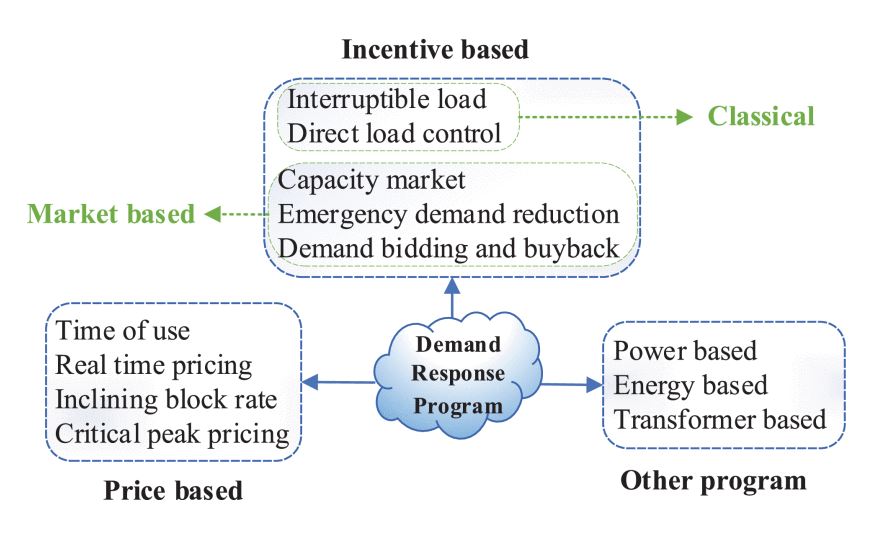} 
\caption{Classification of demand response programs\cite{hussain2021multi}}
\label{im1}
\end{figure}


The time cost (TOU) \cite{datchanamoorthy2011optimal}, immediate pricing (RTP) \cite{el2020novel}, crucial peak costing \cite{herter2007residential}, and tending block rate \cite{borenstein2008equity} are examples of price-based techniques, as seen in Fig. \ref{im1}. The market-based or traditional incentive-based approaches are further subdivided into capacity markets, demands bidding, interruptible loads, direct load control \cite{ericson2009direct}, and urgent demand reduction. 

The International Energy Agency (IEA) predicts that between 2017 and 2040, the world's energy demand will rise by 30\% \cite{hashmi2021internet}. The rationale is that future electric vehicle demand will increase the need for electricity, in addition to new household appliances. Investigating and incorporating renewable energy sources into the current electrical grid will help meet this increasing demand. The primary theme that the authors of the research \cite{hashmi2021internet} address is consumer power usage that is inefficient and ignorant, which results in energy waste and economic losses, especially in developing nations. The creation and application of an Internet of Things (IoT) and cloud computing-based Energy Management System (EMS) for smart grids is the paper's primary focus. The authors have made significant contributions in the form of developing a layered architecture framework for effective energy measurement and control, a reasonably priced platform for monitoring electricity use in real time, and an assessment of the prototype system's performance.

\begin{figure*}[hbt!]
\centering
\includegraphics[height=5.5cm]{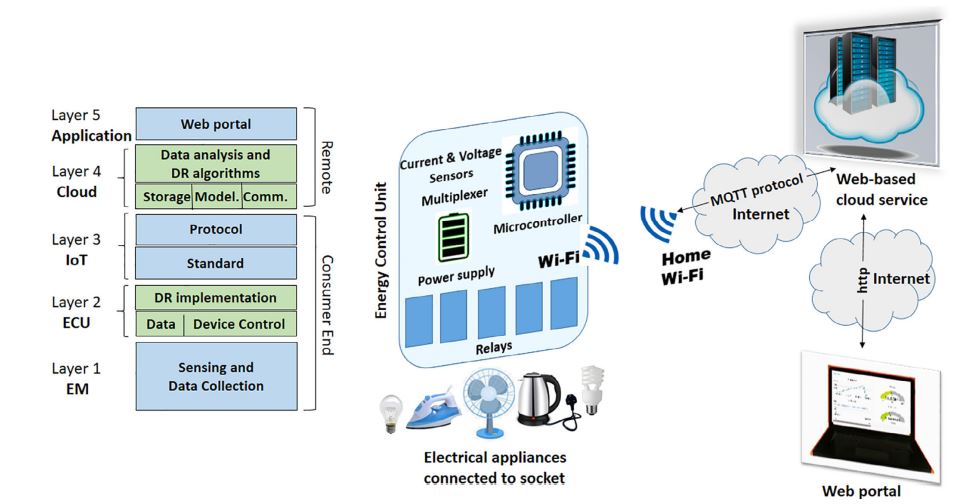}
\caption{(A) The EMS's layered architectural foundation. (B) The EMS system model \cite{hussain2021multi}}
\label{im2}
\end{figure*}

The figure \ref{im2} A displays the layered framework. The layers are defined as follows in a top-down mode:
 1. Utilization, 2. Energy control unit (ECU) 3. Internet of Things (IoT)-based connectivity 4. Cloud analytics 5. Energy monitoring (EM). EMS's system model is displayed in Figure \ref{im2} B. Through the use of voltage and current sensors, whose output is multiplexed before being supplied to the microcontroller's analog input, the model demonstrates electromagnetic fields. 

 Again, using clever pricing and packing techniques, the research \cite{zhao2020smart} presents a Privacy-preserving Data Aggregation (PDA) approach that addresses privacy and security issues in fog-based smart grid interactions. The authors have made significant contributions in the form of developing a multifunctional PDA scheme for challenging statistical functions, introducing smart pricing strategies to guide dynamic energy usage, inventing a new packing technique that can minimize encrypted data size and improve secure computation performance, and using Somewhat Homomorphic Encryption (SHE) to reduce computation and communication overheads. The suggested techniques improve smart grid systems' adaptability, security, and efficiency.

 \begin{figure}
\centering
\includegraphics[height= 5 cm] {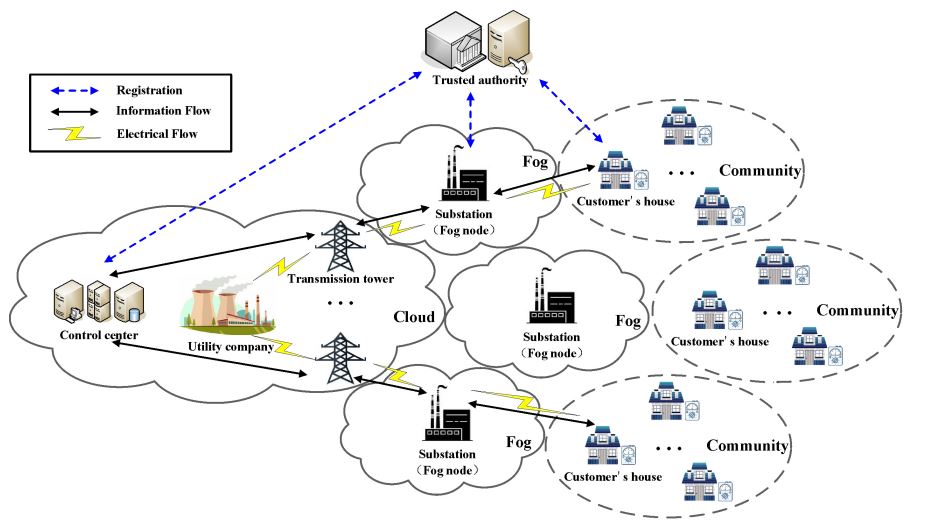}
\caption{Proposed fog-based smart grid's system model of research \cite{zhao2020smart}}
\label{im3}
\end{figure}

Authors \cite{zhao2020smart} depict their system model in Figure \ref{im3}. A trusted authority (TA), a utility, a community college (CC), a few substations, a few fog nodes, and several communities—each with a number of smart meter-equipped customers—make up the fog-based smart grid system.

\subsubsection{Advancements in Load Forecasting Using Machine Learning}
A technology that makes it simple and quick to analyze appropriate judgments for the smart grid's seamless operation is machine learning \cite{umapathy2023machine}. Methodologies like Artificial Intelligence (AI) and Machine Learning (ML) are essential for assessing power consumption, performance, communication path design, and implementation processes in order to optimize \cite{umapathy2023machine}. 
The majority of load forecasting research uses point forecasts based on aggregate system-level data from three different categories: hybrid models, statistical methods, and machine learning techniques \cite{yang2019bayesian}. The time series data is more predictable than local-level data because of its high regularity, which is a result of aggregation. The main challenge that the authors address in this research \cite{yang2019bayesian} is how the irregular and unpredictable data from smart meters makes it more difficult to accurately forecast loads using probabilistic modeling, which is essential for dependable and energy-efficient smart grids. The algorithm \ref{alg:bldr_pld_frcst} is their \cite{yang2019bayesian} claimed Bayesian deep learning algorithm. The utilization of Bayesian deep learning techniques to enhance probabilistic forecasting of load by skillfully managing the variability and unpredictability in power demand is the paper's central focus. The authors' main contributions are as follows: they have developed a clustering-based pooling method to improve data diversity and volume and prevent overfitting; they have proposed a novel multitask Bayesian deep learning algorithm to quantify shared uncertainties across different customer groups; and they have used case studies utilizing public smart meter datasets from the Australian SGSC and Irish CER projects to demonstrate the framework's superior predictive performance. A possible method to formulate the approximate predictive distribution is as follows: 
\begin{equation}
p(y^*|x^*) \approx \int p(y^*|x^*, \omega) q(\omega) \, d\omega. \label{eq:predictive_distribution}
\end{equation}
After that, inference can be carried out by approximating \eqref{eq:predictive_distribution} with a Monte Carlo integral by carrying out \( T \) stochastic forward passes through the network at test time. This is equivalent to determining \( q(\omega) \) by minimizing the objective function. The average of the model's output is then used to estimate the predicted mean.
\begin{equation}
\mathbb{E}_{p(y^*|x^*)}(y^*) \approx \frac{1}{T} \sum_{t=1}^{T} f_{\omega_t}(x^*). \label{eq:predictive_mean}
\end{equation}
The following \ref{im4} is their claimed Bayesian deep learning algorithm. 


\begin{algorithm}
\caption{Bayesian Deep Learning for Probabilistic Load Forecasting}
\label{alg:bldr_pld_frcst}
\begin{algorithmic}
\REQUIRE Sets of load profiles $L_{tr}$, $L_{val}$, $L_{no}$, and $L_{te}$ for training, validation, noise estimation, and testing; number of clusters $K$; number of epochs $E$; stochastic forward passes $T$

\STATE \textbf{Initialization Stage (based on $L_{tr}$)}:
\STATE Compute the normalized representative weekly load profile $L_{i}$ for each customer $i$;
\STATE Cluster the customers into $K$ groups;
\STATE Construct the load profile pool $L_{pool}^{k}$ for each group $k$;

\STATE \textbf{Multitask Learning Stage (based on $L_{tr}$ and $L_{val}$)}:
\STATE Build the multitask Bayesian deep learning model;
\FOR{$e = 1$ to $E$} 
    \FOR{$k = 1$ to $K$} 
        \STATE Sample batches of data points $B^{k}$ randomly from profile pool $L_{pool}^{k}$;
        \STATE Minimize the RMSE loss $\mathcal{L}^{k}$ on $B^{k}$;
    \ENDFOR
    \STATE Evaluate the RMSE and MAE on the validation set $L_{val}$;
    \IF{the validation performance is not improved}
        \STATE Stop
    \ENDIF
\ENDFOR

\STATE \textbf{Forecasting stage (based on $L_{no}$ and $L_{te}$)}:
\FOR{$k = 1$ to $K$}
    \STATE Estimate the inherent noise given by (11);
    \STATE Compute the mean and variance according to (9) and (10), respectively, and evaluate the predictive performance;
\ENDFOR
\end{algorithmic}
\end{algorithm}

Ensuring cybersecurity in smart grids (SG) is essential and has become a critical focus in recent years. The use of artificial intelligence (AI), especially deep learning (DL), offers significant potential for improving SG cybersecurity \cite{berghout2022machine}. In contemporary Smart Grids (SGs), driven by sophisticated computing and networking technologies, condition monitoring depends on secure cyber-physical connectivity \cite{berghout2022machine}. Consequently, some transmitted data, which includes sensitive information, must be safeguarded against various cyber threats due to its vulnerability. The goal of the authors' review \cite{berghout2022machine} is to examine how machine learning (ML) techniques are used in Smart Grids (SGs) to identify and mitigate cyberattacks. They address various instructional paradigms (supervised, unsupervised, and reinforcement learning) and modeling architectures (ordinary, hybrid, and ensemble), and they offer guidelines on ML model selection. They also categorize ML tools based on model complexity and security attributes (Confidentiality, Integrity, and Availability – CIA). The review also outlines the difficulties, limitations, and possible solutions for using ML in SG cybersecurity. The main goals are to provide a thorough review of the most recent machine learning methods used in security governance (SG) contexts for cyberattack analysis and to provide best practices for developing models of prediction for attack detection.

In another research \cite{siniosoglou2021unified}, the authors' primary concern is the serious hazards to cybersecurity and privacy in the networked and diverse Smart Grid (SG) systems. The primary focus of the study is the creation of MENSA, an intrusion detection system (IDS) that classifies cyberattacks on DNP3 and Modbus/TCP protocols by using an Autoencoder-Generative Adversarial Network (GAN) architecture to detect operational irregularities. A high-performance anomaly detection and classification model, the ability to identify a variety of cyberattacks, the validation of the IDS using actual data from different SG environments, and the demonstration of its superior performance over other ML and DL techniques are among the main contributions. In order to create a cohesive DNN framework for (a) anomaly detection and (b) anomaly classification, MENSA integrates the DNNs previously stated.  Given that the Generator-Decoder has been trained to generate normal samples, the likelihood that the actual sample is abnormal increases with the Adversarial Loss. The Adversarial Loss is expressed in the following equation.
\begin{equation}
\text{AdvL}(d_r, d_p) = \|d_r - d_p\|
\label{eq:adv_loss}
\end{equation}

Here,  \(d_r\) and \(d_p\) are the latent model's predictions in the real and generated samples, respectively, and AdvL(x) is the adversarial loss score. Thirteen levels make up the framework of the Generator-Decoder: a layer for input, an output T andh layer, and a series of layers including Dense, ReLU, LeakyReLU, Batch, Normalization, and Dropout layers.\begin{equation}
\tanh(x) = 2s(2x) - 1, \quad \tanh \rightarrow [-1, 1]
\label{eq:tanh_function}
\end{equation}

Where the \(\tanh\) function is described by equation \eqref{eq:tanh_function}. The sigmoid function is \(s(x)\), the output of the \(\tanh\) function is \(\tanh(x)\), and the input vector is \(x\).

\begin{figure}
\centering
\includegraphics[height= 6 cm] {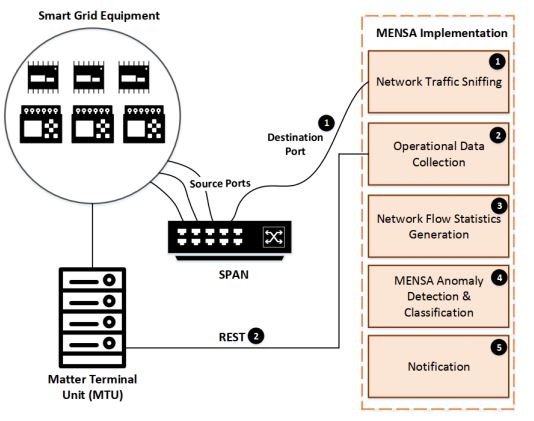}
\caption{ Proposed method's \cite{siniosoglou2021unified} implementation of MENSA. }
\label{im5}
\end{figure}

Fig. \ref{im5} illustrates the implementation of MENSA in an SG context based on the previously given observations. MENSA operates independently of software sensors and services within the SG environment, utilizing a specialized computing system. Therefore, it has no effect on the computational power or the SG equipment's ability to function normally. 
\section{Comprehensive Analysis of Smart Grid Components, Technological Advancements, and Their Role in Renewable Energy Integration}
Smart grids offer a myriad of advantages, including improved reliability, increased efficiency, and enhanced ability to incorporate renewable energy sources \cite{kataray2023integration}. They enable real-time monitoring and management of energy flows, leading to optimized performance and reduced energy losses \cite{kataray2023integration}. Recent technological advancements have significantly enhanced the capabilities of smart grids. This section will explain the components of Smart Grid, technological advancement in this field and how smart grids are a cornerstone for the effective integration of renewable energy sources into the power system. 

\subsection{Components of Smart Grid}

\begin{figure}
\centering
\includegraphics[height= 4.5 cm] {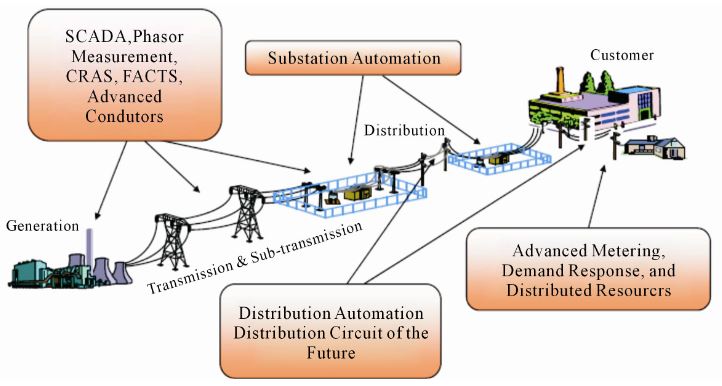}
\caption{Elements of a Smart Grid \cite{vijayapriya2011smart}. }
\label{im6}
\end{figure}

A smart energy system is an electricity network that can automatically and sensitively integrate the behavior of all users linked to it, encompassing generators, clients, and the individuals who do both, in order to efficiently offer sustainable, inexpensive, and secure electrical supplies. In the research paper \cite{vijayapriya2011smart} authors describe the componests of smart grid through Figure \ref{im6}. The authors of the research paper \cite{chong2020review} explain that when using a smart grid system, this typically includes. Customer Data Generation, Transfer, Distribution, and Interaction. 

\begin{enumerate}
        \item \textbf{Custome:}
        Building automation and control systems serve as the central management hub of a building. These systems integrate instrumentation, control, and management technologies for all structures, plants, outdoor facilities, and equipment within the building.
        
        \item \textbf{Bulk Generation:}
        Smart generation involves increased use of power electronics to manage harmonics, address fluctuating generation failures from renewable energy, and enhance the flexibility of fossil power plants due to the higher integration of renewable energy sources.
        
        \item \textbf{Transmission and Distribution:}
        The energy management system (EMS) operates as the control center for the transmission network \cite{kabeyi2023smart}, requiring open architecture for easy IT integration and better backups to prevent blackouts. Power electronics, including HVDC and FACTS systems \cite{kabeyi2023smart}, enable control of power flow and help increase transport capacity without causing short-circuit increases.
        
        The Distribution Management System (DMS) controls power grids, and in regions prone to outages, the Outage Management System (OMS) is critical \cite{solomon2022outage}. Other essential components include failure location interfaces on Geographic Information Systems (GIS) \cite{ukoba2023geographic} and smart meters with communication links. Advanced Metering Infrastructure (AMI) allows for remote measurement configurations \cite{phing2024brief}, dynamic rates, power quality monitoring, and load control. More sophisticated systems combine measurement infrastructure with distribution automation.

        \item \textbf{Communication:}
        Current data communication within the network is primarily one-way, leading to longer recovery periods during faults. Many devices, such as transformers, remain isolated and cannot be monitored online. Installed protection equipment is largely electromechanical and lacks integration with digital sensors, limiting the information available to operators in control rooms.
        
        To enhance the intelligence of the power grid, infrastructure upgrades are necessary. These upgrades include adding digital sensors, auto protection systems, online monitoring, wireless sensor networks, measuring instruments, CCTV cameras, and other smart devices \cite{chen2023control}. The adoption of smart grid technology, coupled with the Internet of Things (IoT), will enable real-time data interaction and two-way communication \cite{alomar2023iot}. Additionally, Multi-Agent Systems (MAS) are highlighted in literature for their potential positive impact on smart grid development \cite{sajid2024multi}. MAS allows for autonomous resource management through agent-based modeling and simulation, offering flexibility for resources to collaborate, coordinate, and negotiate to achieve goals efficiently \cite{yarahmadi2024improving}. Intelligent agents can bring any planned smart grid feature into reality by acting as distributed control building blocks.

\end{enumerate}

\subsection{Integration of Renewable Energy Sources into Smart Grid}

It is difficult to integrate a large percentage of power through non-dispatchable Renewable Energy Sources into a power supply system \cite{panda2023recent}. Installing storage devices to offset the variations in power generation is one alternative taken into consideration in numerous studies pertaining to potential power systems. It is unclear, therefore, whether soon storage devices would be required and how the technique of integration will be impacted by various storage settings. The authors of the research \cite{panda2023recent} provide a modeling approach to study the impact of capacity and efficiency on the journey towards a 100\% RES scenario using for a long time solar and wind energy energy production data series. Using data for Germany, they applied their method and discovered that, if enough flexible power plants are used to supply the remaining energy, an optimal mix of solar and wind resources may meet up to 50\% of the country's total electricity demand without the need for curtailment or storage devices. 

The seamless incorporation of renewable energy resources within current power systems is critical to the global shift towards sustainable energy solutions \cite{etukudoh2024electrical}. This paradigm transition is a response to the pressing need to mitigate climate change, cut emissions of greenhouse gases, and guarantee a sustainable and ecologically friendly energy future. A viable route to achieving these objectives is through the use of renewable energy sources, such as wind, hydroelectricity, solar energy, and geothermal power \cite{kataray2023integration}. The present study delves into the intricate process of incorporating renewable energy sources into power grids, emphasizing the key obstacles and prospects that underpin this revolution \cite{barman2023renewable}. The intrinsic unpredictability and intermittent nature of renewable energy present technological difficulties that call for creative fixes, such as grid upgrading and cutting-edge storage technology. This integration requires advanced technologies and strategies to ensure reliability, efficiency, and stability of the power supply. Key aspects include: Advanced Metering Infrastructure (AMI), Distributed Energy Resources (DERs), Smart Sensors and Actuators, Communication Networks, Energy Storage Systems, Advanced Control Systems, Grid Flexibility, Predictive Analytics and Machine Learning, and Regulatory and Policy Support. 

Systems such as Advanced Metering Infrastructure (AMI) are essential for delivering precise and up-to-date data on electricity usage \cite{yadav2024smart}. Accurate billing is made possible by these systems, which also improve energy management for utilities and customers alike. Renewable energy sources like solar and wind turbines, as well as energy storage devices, are examples of distributed energy resources, or DERs \cite{yadav2024smart}. Decentralized power generation and a decrease in the use of fossil fuels are made possible by DERs \cite{shaukat2023decentralized}. Smart grids may increase resilience and sustainability by utilizing regional energy generation and storage through the integration of DERs \cite{shaukat2023decentralized}. Renewable energy sources like solar and wind turbines, as well as energy storage devices, are examples of distributed energy resources, or DERs. Smart grids may increase resilience and sustainability by utilizing local energy generation and storage through the integration of DERs.

Data collection on different grid characteristics is greatly aided by the use of intelligent sensors and actuators. These gadgets enable automatic control operations that preserve the efficiency and stability of the grid. Data collection on different grid characteristics is greatly aided by the use of intelligent sensors and actuators. These gadgets enable automatic control operations that preserve the efficiency and stability of the grid. In order to balance the variations in power generation from non-dispatchable RES, energy storage technologies are essential. Sophisticated control systems oversee the smart grid's integration of diverse energy sources. In order to ensure that energy is used effectively and efficiently, these systems optimize energy flow and distribution in real-time. Integrating flexible power plants into the grid entails promptly adjusting output to balance supply and demand. Machine learning and predictive analytics are effective methods for predicting patterns in energy production and consumption. Because these technologies offer precise forecasts and insights, proactive management of grid operations and resources is made possible. Support from laws and policies is necessary to promote the use of renewable energy technology. In order to encourage grid modernization and integration, this involves setting standards, constructing frameworks, and offering incentives.

\section{Challenges for Smart Grids}

The development of smart grids (SGs) is characterized by several essential elements, such as improved power systems, communication and standardization, computational intelligence, consideration of environmental and economic issues, and test beds \cite{moreno2021comprehensive}. Human ability, governmental policies, technology, and electrical issues are a few of the obstacles to sustainable growth. By incorporating power system enhancement techniques, improving interaction and standards, establishing computational intelligence equipment that supports the environmental, financial, and development of clean energy, and demonstrating a suitable testbed that demonstrates the benefits of SG \cite{goudarzi2022survey}, these features make it easier to develop an energy-efficient system that meets current demand. 

The computational intelligence component includes sophisticated analytical tools that will use heuristics, evolution programming, tools for decision support, and adaptive optimization techniques to optimize the bulk power network. These tools are promising for the SG's design and computing needs. The growth of renewable energy makes use of variability, and it is both technically and financially feasible to exploit it to fulfill demand uncertainty shortages, boost reliability, lower losses, and lessen the carbon footprint left by gas and thermal energy sources \cite{osti_1804705}. Automation, standards, and communication must be developed in order to guarantee quick decision-making that encourages effectiveness and responsible operation. Customers and the utility make these decisions \cite{chen2023control}. By creating and implementing the policies and guidelines for administering, running, and promoting the SG networks, security and interoperability concerns are also ensured \cite{chen2023control}.

Technological and infrastructure challenges present significant hurdles to the implementation of smart grids. Many existing power grid infrastructures are outdated and not equipped to handle the advanced technologies required for smart grid operations, necessitating substantial investment and time for upgrades\cite{pandiyan2023technological}. Additionally, ensuring interoperability among various components and technologies from different vendors is crucial for the smooth functioning of a smart grid \cite{pandiyan2023technological}. This requires meticulous coordination and standardization to ensure seamless integration and operation of the diverse systems involved.

There is a significant need for a skilled workforce capable of managing and operating smart grid technologies \cite{ezeigweneme2024smart}. To bridge this skill gap, continuous training and development programs are essential, ensuring that personnel are adequately equipped to handle the complexities of smart grid systems \cite{ezeigweneme2024smart}.

The increased use of wireless communication technologies in smart grids raises concerns about potential health impacts due to electromagnetic radiation \cite{maruf2020adaptation}. Additionally, the sustainability of materials used in the production and disposal of smart grid components needs to be managed carefully to minimize environmental impact \cite{maruf2020adaptation}. Addressing these concerns is crucial for ensuring the long-term viability and public acceptance of smart grid technologies.

So, there are several factors which are throwing challenges for implementation smart grid. In this section we will discuss elaborately about three major challenges below.

\subsection{Intermittent Nature of Renewable Energy}

In order to provide the electrical grid with the features and capabilities required for a seamless transition towards sustainability and the integration of renewable energy sources, the idea of the intelligent grid was developed \cite{khalid2024smart}.  As increasing amounts of sources of clean energy are incorporated into the electrical system and the market for electricity gradually shifts from a centrally controlled to a more dispersed structure, the requirement for SG grows tremendously. This review paper's \cite{khalid2024smart} goal is to draw attention to the relevant SG challenges that must be overcome in order for the technology to become more widely accepted from the standpoint of user acceptance and versatility for energy system organizers and operators with regard to regulatory and serviceability requirements. Table \ref{tab1} lists a number of standards, reviews, and benchmark descriptions related to the components of systems for smart grids.

\begin{table}
\caption{Assessments of the most recent developments in technology, security, standards, and risks in Singapore.}
\label{table}
\setlength{\tabcolsep}{3pt}
\begin{tabular}{|p{22pt}|p{155pt}|p{40pt}|}
\hline
Ref.& 
Elucidation & 
Publication Year \\
\hline
\cite{bou2013communication} & 
HAN and NAN security measures and communication technology kinds are reviewed and categorized. & 2013 \\

\cite{he2016cyber} & 
Categorization and remedies for cyberattacks targeting SG's operating technology & 2016 \\

\cite{leiva2016smart} & 
An overview of global smart metering regulations and its framework & 2016 \\

\cite{anzalchi2015survey}, \cite{wei2018review} & Threat classification, types, and mitigations for advanced metering infrastructure security. 
 & 
2015, 2018 \\

\cite{wang2011analysis}, \cite{leszczyna2018cybersecurity} & 
Several SG security standards in IEC/ISO, the International Electrical and Electronic GB, and NIST are reviewed and summarized. a thorough explanation of each standard's definition is included in the survey of the evaluation and selection criteria. & 
2011, 2018 \\

\cite{gunduz2020cyber} & 
An overview of the risks to cyber security, a categorization for network layers, and an update on relevant research along with preventative actions. & 2020 \\

\cite{srivastava2021emerging} & 
Detailed explanation of security attacks in the Internet of Things and operational technologies, along with countermeasures. & 
2021 \\

\cite{rajendran2021comprehensive} & 
standardization of EVs and charging stations along with a thorough explanation of the most recent advancements and administration techniques. & 2021 \\

\cite{liberati2021review} & 
Technical categorization of cyber-physical assaults in SG using a mathematical methodology for state estimation and fake data injection. & 2021 \\

\cite{hasan2023review} & 
Guidelines, norms, suggestions, and limitations of cyber-physical systems and cyber-security in SG. & 
2022 \\

\cite{kim2023smart} & 
The innovative metering structures, technological innovations, and operational technologies are all covered in this concise, clear description of SG security, along with the most recent findings on countermeasures. & 
2022 \\

\hline
\end{tabular}
\label{tab1}
\end{table}

The problems with optimal flow of energy are related to the efficient distribution and transmission of electricity. Unlike other obstacles, transience, modularity potential, and topological dependency are only a few of the many diverse reasons why RE sources give birth to related issues in the case of optimal power flow challenges. The main concerns related to stability issues include controlling the power system's voltage and frequency as well as recovering the system after outages.

\subsection{Cybersecurity Concerns}

The communication networks and computer platforms that run and oversee the entire grid are starting to worry a lot about cyber security in the context of smart grid systems. It is difficult to implement consistent security measures across all of the network segments due to the characteristics of smart grid networks, including heterogeneity, scalability, bandwidth, and delay limits. Consequently, further study is needed to create standards and methods that may affordably satisfy the needs of the smart grid network \cite{7454291}.

Many studies have been conducted recently to guarantee the accuracy of cyberattack detection and identification, as well as to reduce computational time and enhance resilience against external influences \cite{mohammadi2021emerging}. Different obstacles must be overcome by power systems operators in order to detect and identify cyberattacks in the smart grid's infrastructure \cite{mohammadi2021emerging}. The following are the main obstacles:
\begin{itemize}

\item The precision with which cyberattacks are identified and detected.
\item Computational load during the detection and classification of cyberattacks.
\item Robustness against a range of variables influencing the identification and detection of cyberattacks.
 \end{itemize}
 
These dangers could represent serious threats to people's privacy, including the possibility of sensitive customer data being stolen or the company closing its doors permanently \cite{faquir2021cybersecurity}. These hazards concern consumers not only when they use the internet but additionally when they are at home, where attackers may be able to obtain personal data.  Given how simple it is to carry out, phishing may be the initial step in endangering clients and businesses. With the help of social engineering, hackers could obtain vital information about an organization—in this case, the electricity supplier—by using data from consumers who have not disposed of their bills or payment receipts in a responsible manner \cite{faquir2021cybersecurity}.  
 
Denial-of-Service (DoS) assaults are strategic in nature, and they encompass any attack directed towards availability \cite{ortega2023review}. The availability of the top services to smart grids implies that there is a possibility of a denial-of-service assault against the smart grid. Secure and dependable connectivity is required for the Smart Grid. Because the Smart Grid uses distributed architectural systems to spread connections to numerous devices over a greater region, the necessity for connectivity must be dependable and safe \cite{ortega2023review}. Should a DoS attack transpire upon the Smart Grid, it will incur significant losses. 
 
The main worry is that malware spreading might pose a significant risk to the Smart Grid. Malware that can be used to infect both devices and organization servers can be created by the attackers \cite{akhtar2018malware}. Through the use of malware propagation, an attacker can alter the way that devices or systems operate, giving them access to obtain sensitive data \cite{akhtar2018malware}.
 
Spoofing attacks include things like traffic monitoring and eavesdropping. By keeping an eye on network traffic, the attacker can obtain sensitive data. This danger will be faced by the Smart Grid \cite{ding2022cyber} because to its extensive network, which consists of numerous network nodes and is difficult to maintain for the devices linked to the main network. The main worry in protecting data worldwide is that the Smart Grid presents the highest danger of data theft \cite{ding2022cyber}.

\begin{figure}
\centering
\includegraphics[height= 4.5 cm] {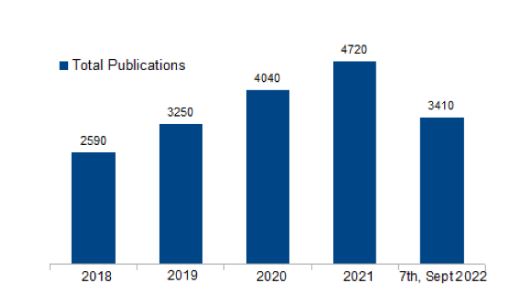}
\caption{Papers sorted by year found on Google Scholar using the search terms "smart grid" AND "cyber threats" OR "cyberattacks" OR "vulnerabilities." \cite{ding2022cyber}. }
\label{im7}
\end{figure}

As shown in Figure \ref{im7}, smart grid research has drawn the attention of numerous academics in the past, and the number of publications has grown exponentially. To find the precise studies, we used the conjunction (AND) as well as disjunction (OR) rules to search for terms like "smart grid," "cyber threats," "cyberattacks," and "vulnerabilities."

\subsection{Integration of Electric
Vehicles}

One of the main contributors to air pollution and climate change are emissions from internal combustion engine (in short ICE) cars. Compared to these internal combustion engine (ICE) vehicles, electric vehicles (in short EVs) are starting to make more sense \cite{patil2020grid}. Since large-scale production and recent advancements in battery technology, the cost of EVs has decreased. The current electrical power system (in short EPS) will be under tremendous strain in the near future due to the widespread adoption of EVs \cite{patil2020grid}. When EVs are scheduled optimally, the strain on the current network can be minimized and large-scale EV integration can be supported. Several energy market participants may profit monetarily from the incorporation of these EVs \cite{patil2020grid}. 

To examine the cost savings under the three charging operation modes—random charging, regulated charging, and V2G charging—an electrical power cost model is set up \cite{wu2021benefits}. The findings show that EV adoption can successfully raise the load factor of the power grid and lower the cost of electricity delivery. The marginal cost of electricity under the unpredictable charging mode is 556.5 CNY/MWh \cite{wu2021benefits}, that is less than the power supply expense of 597.1 yuan/MWh for the grid. Without V2G, controlled charging can further flatten the load curve and reduce the cost of the power supply to 495.2 CNY/MWh \cite{wu2021benefits}. The best overall revenue under the V2G model might be 148.4 million CNY, although this is complicated by the need to account for battery deterioration expenses. 

The electricity grid has a chance to lower operating costs and increase load factor with the advent of electric vehicles. However, the primary obstacles are institutional and economic in nature. This is primarily because there are no legal frameworks in place for assessing flexibility at the distribution level, which creates confusion about the worth of these flexibility offerings.
 
The first difficulty in integrating electric vehicles (in short EVs) with smart grids is the intricate environment distribution system operators must navigate. The institutional and financial restrictions pertaining to the value of the flexibility services offered by EVs are one of the main obstacles. The process of integration is significantly impacted by technical issues as well \cite{venegas2021active}. Utilizing EV flexibility can be done in a variety of ways, depending on how the electrical grid is arranged spatially. These tactics can be user-centered (behind-the-meter) or system-wide (transmission threshold, wholesale markets). An understanding of the legislative and economic landscape is essential for a successful EV integration \cite{venegas2021active}. Policies and regulations that are always changing must promote flexibility bargaining at the distribution stage from both EVs and other demand-side mechanisms. User-related issues present yet another big obstacle. End-user behavior—specifically, how EVs are utilized and charged—and acceptance of control tactics are key factors in the achievement of flexibility services \cite{venegas2021active}. All things considered, EV integration into smart grids necessitates resolving technological, financial, legal, and user-related issues.

\section{AI and Data Analytics in Smart Grid}

\begin{figure}
\centering
\includegraphics[height= 4.5 cm] {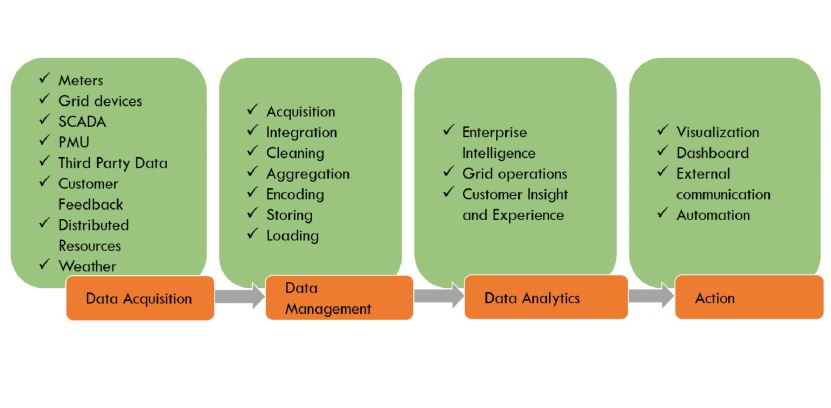}
\caption{High-level view of the flow of data into the utility \cite{syed2020smart}. }
\label{im8}
\end{figure}

The success of the big data analytics component will determine if the smart grid power paradigm succeeds. This covers the efficient collection, transfer, processing, interpretation, visualization, and application of large data. In addition to discussing the analysis of big data in relation to the smart grid, the paper \cite{syed2020smart} offers in-depth insights on a number of big data technologies. The problems and opportunities posed by the emergence of big data from electrical grids and machine learning are also covered in the \cite{syed2020smart} study. Figure \ref{im8} shows an executive summary of the data stream into the utility. 

\begin{figure}
\centering
\includegraphics[height= 6 cm] {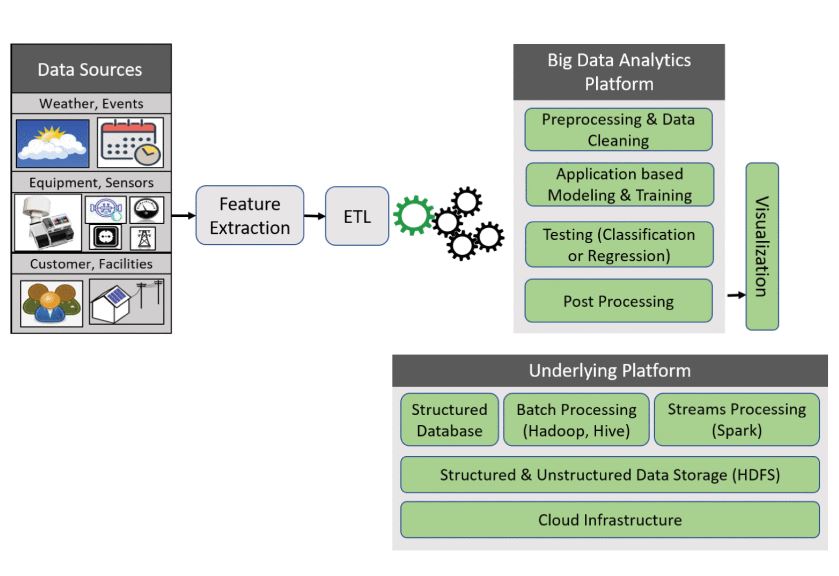}
\caption{Process of big data analytics. \cite{syed2020smart}. }
\label{im9}
\end{figure}

Pre-established methodologies are necessary for Big Data analytics due to the large amount of data \cite{syed2020smart}. The speed and diversity of data can present difficulties for data analytics procedure. Real-time processing of the smart grid's data is essential because it allows for the identification of important trends that can be used to inform decision-making \cite{syed2020smart}. The process of identifying patterns and actionable information in the available data is known as data analytics. Figure \ref{im9} shows big data analysis procedure.

The following categories can be used to broadly group the AI methods used in smart grid systems \cite{omitaomu2021artificial}.
\begin{itemize}
 
\item ES: Human loop expert utilized to solve specific challenges.
\item Supervised learning: an artificially intelligent paradigm where the outputs of fresh inputs are predicted by studying the mapping of the inputs to the outputs.
\item Unsupervised learning:  ML course in which the similarities and differences between the data are determined by using the unlabeled data.
\item Reinforcement learning (RL): Its intelligent agents approach, which tries to optimize the idea of cumulative reward, sets it apart from supervised and unsupervised learning.
\item Ensemble methods: To get around one algorithm's drawbacks and improve overall performance, combine the output of multiple AI systems.
\end{itemize} 

\begin{figure*}[hbt!]
\centering
\includegraphics[height= 15cm]{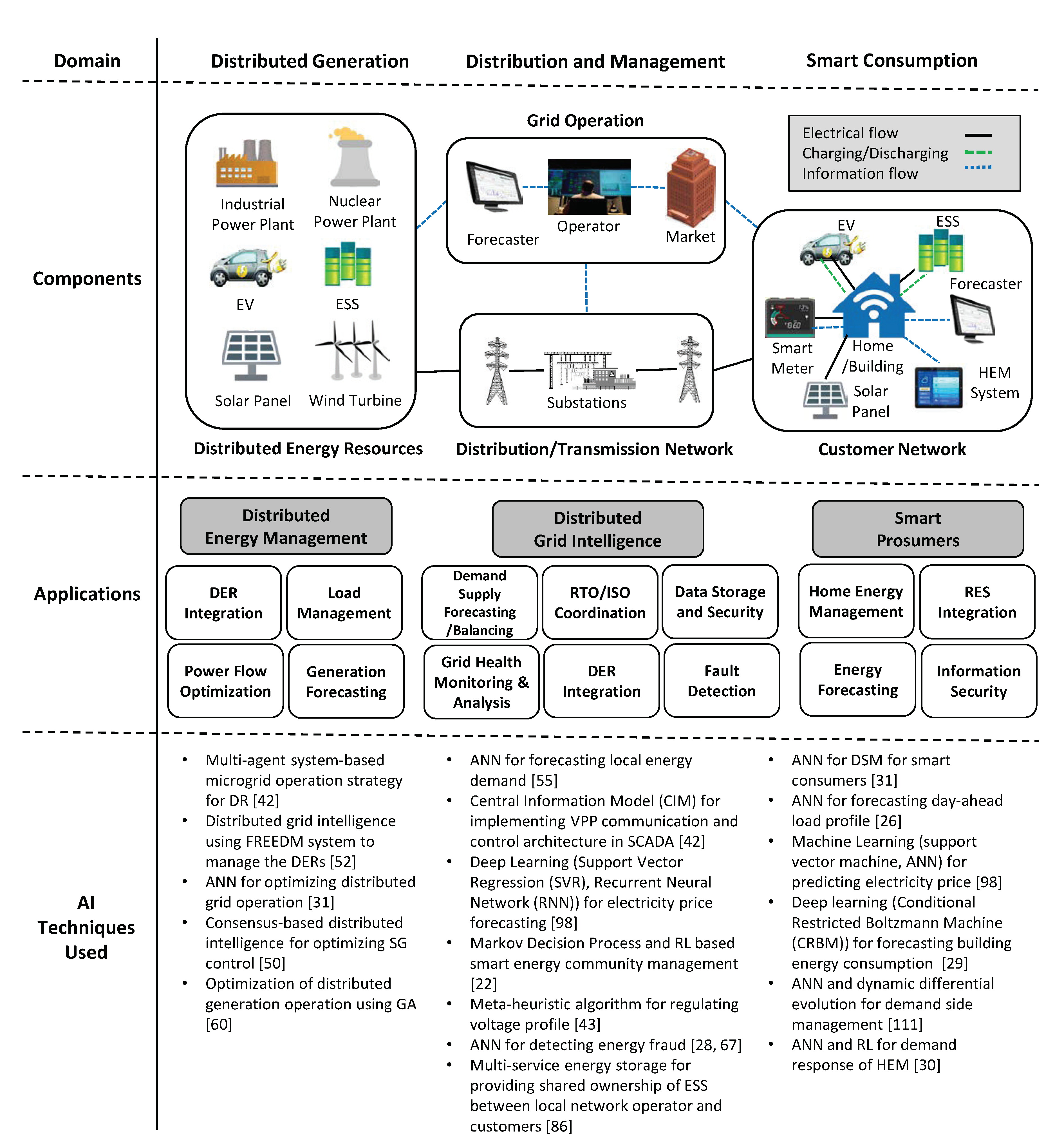}
\caption{Summary of AI methods utilized in decentralized intelligent power grids \cite{ali2020state}}
\label{im ai tech}
\end{figure*}

The use of artificial intelligence (AI) approaches is becoming more popular as they offer strong and promising tools for controlling and analyzing stability in smart grids. The research paper's authors \cite{shi2020artificial} also list some of the main obstacles that these applications must overcome in the real world: strict data requirements, unequal learning, AI interpretability, challenges with learning transfer, AI resilience for interacting quality, and AI resilience against adversarial examples or attacks. Stability analysis aims to identify the kind of instability \cite{shi2020artificial} and explore the power grid's ability to sustain stability in the face of disruptions. It is only possible to conduct timely emergency control measures if the precise nature of the instability can be ascertained promptly. Figure \ref{im ai tech} illustrates the comprehensive concept of distributed smart grids incorporating AI methods. Utilizing intelligent techniques for optimizing controllable loads leads to cost savings.

As an emerging technology that can create a more intelligent and decentralized energy paradigm while incorporating high intelligence in operational and supervisory decision-making, SG is strongly associated with Deep Learning (DL). Driven by the remarkable achievements of DL-based prediction techniques, the research paper \cite{9395437} aims to offer a comprehensive overview from a wide angle of the state-of-the-art developments of DL in SG systems. First, the approach of this review has been categorized using bibliometric analysis \cite{9395437}. Moreover, authors \cite{9395437} explore the workings of some of the current DL algorithms taxonomistically. The DL technologies that enable in SG, like distributed computing, edge intelligence, and federated learning, are then highlighted by them \cite{9395437}. 

An additional study \cite {9293294} suggests AEBIS, a blockchain-based, AI-enabled electric vehicle incorporation solution, for managing electricity in a framework of smart grid. The EV fleet is used as both a provider and a consumer of electrical power inside a virtual power plant (VPP) platform. The technology is built on federated learning techniques and ANN for EV charge prediction. According to the evaluation results, in the traditional training scenario, the suggested approach \cite {9293294} obtained a substantial power consumption prediction with a R 2 score of 0.938. There was only a 1.7\% loss in accuracy when using a federated learning strategy. 

The integration of machine learning and AI can facilitate the smart grid's ability to make astute decisions and react to unforeseen situations such as power outages, abrupt fluctuations in renewable energy generation, or other catastrophic occurrences \cite{9084590}. Additionally, machine learning can be used to predict equipment breakdowns, forecast energy demand, and power generation from intermittent sources \cite{9084590}. It can also be used to capture customer usage trends. To maintain an equilibrium of power supply and demand, demand management signals can be activated and energy dispatch decisions made with the use of reinforced learning. The smart grid is susceptible to cyber security risks due to its reliance on wireless technologies \cite{9084590}. 

\begin{figure}
\centering
\includegraphics[height= 6 cm] {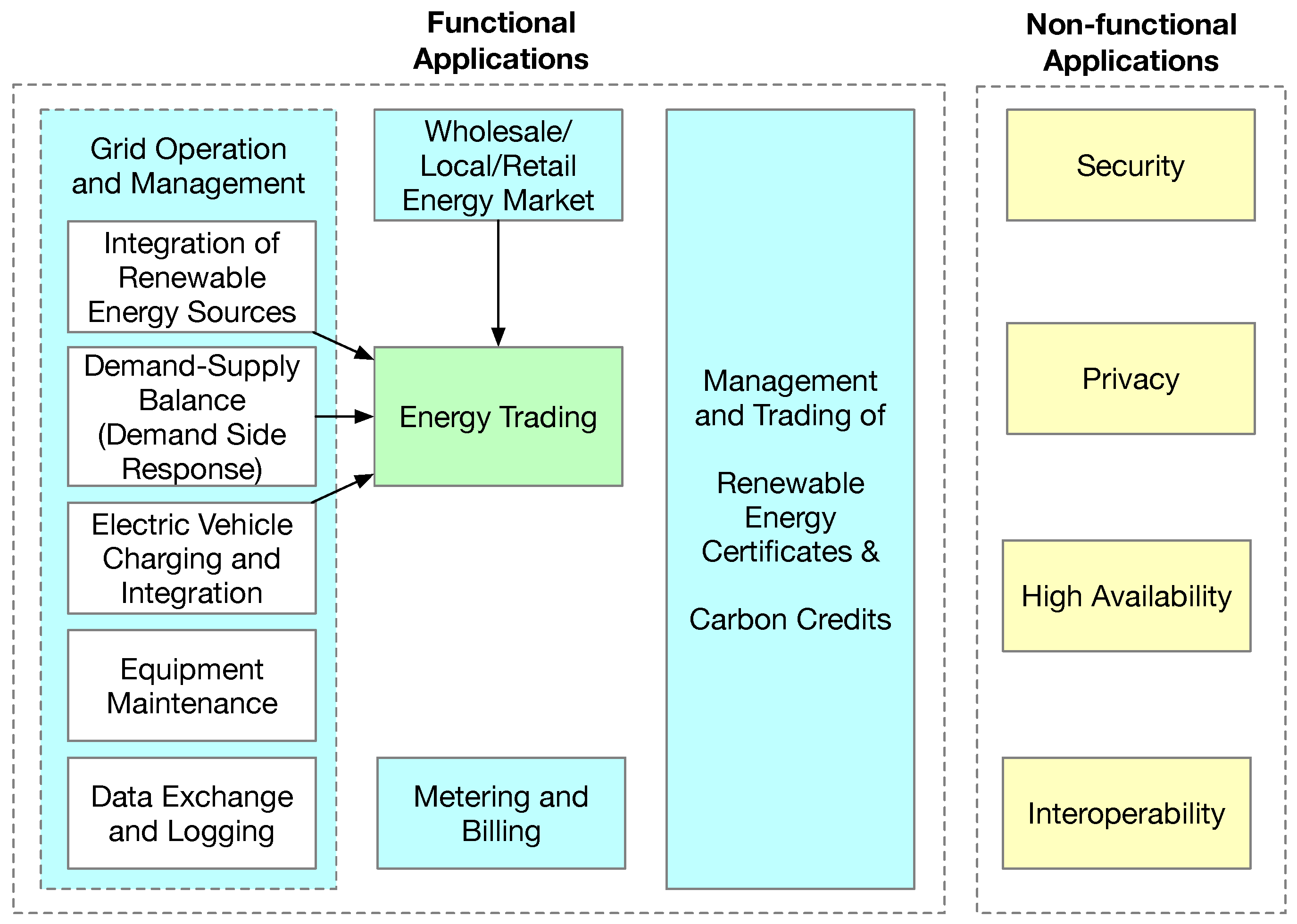}
\caption{Taxonomy of applications for blockchain-based smart grids \cite{zhao2023blockchain}. }
\label{energies}
\end{figure}

Current systematic reviews categorize blockchain applications in smart grids in various manners \cite{zhao2023blockchain}. Authors \cite{zhao2023blockchain} consolidate these reviews and introduce a taxonomy of blockchain applications. Initially, this taxonomy is based on whether the application addresses functional or non-functional requirements, and subsequently on the specific goals of each requirement type, as depicted in Figure \ref{energies}. They suggest a five-level scale to assess maturity levels. The ranking of industry-led projects is conducted based on their own research.

Using a fully examined typical Irish house, the article \cite{pallonetto2019demand} evaluates control algorithms enabling demand response techniques in the residential sector. A calibrated structure simulation framework was created by them \cite{pallonetto2019demand} to assess these tactics with varying time-of-use tariffs. On a combined heat pump and thermal storage system, two algorithms—a rule-based method as well as a predictive ML approach—were investigated. The predictive algorithm demonstrated notable decreases in comparison to a baseline: 37.9\% in carbon emissions, 39\% in utility generation expenses, and 41.8\% in electricity expenditure. The study \cite{pallonetto2019demand} emphasizes how machine learning may be used to optimize home energy use and lower carbon emissions.

To create the optimization and oversight mechanism, precise energy demand estimate and very accurate short- and/or long-term forecasting findings are needed. As a result, distributed demand response (DR) initiatives and machine learning (ML) approaches are being used to estimate future energy demand requirements with satisfying outcomes. This research \cite{9765492} aims to provide a comparative examination of machine learning (ML) algorithms using short-term load forecasting (in short STLF) with respect to forecast error and accuracy. The authors \cite{9765492} have determined that, in comparison to other algorithms, the DTC yields somewhat superior outcomes based on their implementation and analysis. In order to fine-tune the control variables, they \cite{9765492} developed the enhanced DTC (EDTC) by combining the fitting operation, loss function, and gradient boosting within the DTC mathematical model. The suggested EDTC algorithm yields improved forecast outcomes (i.e., 100\% F1, which is 100\% precision, 99.21\% accuracy in training, and 99.70 percent testing accuracy), according to the implementation findings.

\section{Future Research Scope}

Future study should focus on three important areas, according to the authors of the research paper \cite{el2014smart}, for the purpose of effectively implement the smart grid. For efficiency and a superior power supply, components of the distribution system such as harmonic filters and capacitors must be sized and positioned optimally. For realistic resource dispatching, combining prediction as well as optimization techniques and creating interconnected forecasting suites are essential \cite{el2014smart}. Post-reconfiguration modules are necessary for adaptive reorganization of distribution networks (in short DNs) in order to guarantee optimal outcomes \cite{el2014smart}. For real-time network assessment and adaptability, strategic placement of sensors for estimation of state and observability is essential. Additionally, to properly control the stochastic nature of demand for energy, combining both static and dynamic storing resources—including customer-based storage—is crucial. 
 
Future electrical grids will function autonomously to the extent that real-time measurements are obtained by IoT sensor nodes, which then use edge computing to analyze the measurements in a distributed way. The Energy Management System (EMS) then receives the data and stores it in the cloud. Control algorithms over self-resilient grids would be developed using Artificial Intelligence (AI) for large data analytics, using ML as well as DL approaches \cite{atitallah2020leveraging}, \cite{kotsiopoulos2021machine}. Furthermore, it is evident that the rigid, hierarchical grid topology—which was regulated by monopoly markets—is transitioning to a delegated system and experiencing significant changes in authority and operation \cite{mollah2020blockchain}, \cite{ahl2020exploring}, \cite{ahl2020exploring}. The goal of the thorough analysis \cite{yapa2021survey} is to examine how blockchain technology might be applied to Smart Grid 2.0 to enable a smooth decentralization process. The article \cite{yapa2021survey} also explains the blockchain's function in each scenario and how it will be applied to future smart grid operations. 

A smart grid is an intelligent network of electricity that integrates the actions of all stakeholders, including generators, consumers, and one who does both, in order to supply electricity with effectiveness, environmental responsibility, economy, and security, according to the Strategic Deployment Document for Europe's Electricity Networks of Future \cite{smartgrids2008strategic}. A great deal of research is being done to create smart grids. As to the research article \cite{butt2021recent}, there is ample scope for future investigations into various facets across several domains of smart grids. This includes the fields of microgrid integration, communication, prediction, power flow optimization, scalability, economical factors, demand and energy management systems, compliance with interoperability standards, data encryption, and—above all—automation of generation, transmission, and distribution.Technological and gadget advancements have the potential to transform energy use in a cost-effective and eco-friendly manner. The evolution of the smart grid concept offers the ability to integrate more renewable energy sources and reduce carbon emissions to meet all future energy needs \cite{butt2021recent}. Over the past ten years, there has been a significant growth in research and innovation on smart grids. For this reason, the concept and virtual reality phases of smart grid technology have given way to the implementation phase. There have been 26,668 distinct research publications \cite{butt2021recent} published on smart electricity systems in the previous 10 years, indicating that scholars are quite interested in this topic.

\section{Conclusion}

An overview of the history, features, technology, applications, benefits, possibilities, difficulties, as well as future potential of the smart grid is provided in this paper.  This research also makes a detailed effort to determine the importance of merging renewable energy sources alongside smart grids. 
The major aim of the study is to determine the prime obstacles that smart grids must overcome, including the inconsistent supply of renewable energy, worries about cybersecurity, and the incorporation of electric vehicles under the grid.
 The authors also look into how data analytics and AI might be used to solve these problems and improve smart grid performance.
 The report offers helpful suggestions for future areas of study in this quickly expanding discipline by evaluating the state of the field and emphasizing important concerns. Rejuvenating the power infrastructure makes it possible to implement demand side management plans for global energy management.

 \textbf{Funding Details:} his research is funded in part by NSF Grants No.
2306109, and DOEd Grant P116Z220008 (1). Any opinions,
findings, and conclusions expressed here are those of the
author(s) and do not reflect the views of the sponsor(s).


\bibliographystyle{IEEEtran}
\bibliography{bibtex.bib}

\begin{thebibliography}{100}
\providecommand{\url}[1]{#1}
\csname url@samestyle\endcsname
\providecommand{\newblock}{\relax}
\providecommand{\bibinfo}[2]{#2}
\providecommand{\BIBentrySTDinterwordspacing}{\spaceskip=0pt\relax}
\providecommand{\BIBentryALTinterwordstretchfactor}{4}
\providecommand{\BIBentryALTinterwordspacing}{\spaceskip=\fontdimen2\font plus
\BIBentryALTinterwordstretchfactor\fontdimen3\font minus \fontdimen4\font\relax}
\providecommand{\BIBforeignlanguage}[2]{{%
\expandafter\ifx\csname l@#1\endcsname\relax
\typeout{** WARNING: IEEEtran.bst: No hyphenation pattern has been}%
\typeout{** loaded for the language `#1'. Using the pattern for}%
\typeout{** the default language instead.}%
\else
\language=\csname l@#1\endcsname
\fi
#2}}
\providecommand{\BIBdecl}{\relax}
\BIBdecl

\bibitem{gegner2017phasor}
K.~M. Gegner, ``Phasor measurement unit data visualizations and their role in improving operation of the electric grid,'' Ph.D. dissertation, University of Illinois at Urbana-Champaign, 2017.

\bibitem{nair2018assessment}
V.~Nair, R.~Litjens, and H.~Zhang, ``Assessment of the suitability of nb-iot technology for orm in smart grids,'' in \emph{2018 European Conference on Networks and Communications (EuCNC)}.\hskip 1em plus 0.5em minus 0.4em\relax IEEE, 2018, pp. 418--423.

\bibitem{khan2022energy}
N.~Khan, Z.~Shahid, M.~M. Alam, A.~A. Bakar~Sajak, M.~Mazliham, T.~A. Khan, and S.~S. Ali~Rizvi, ``Energy management systems using smart grids: an exhaustive parametric comprehensive analysis of existing trends, significance, opportunities, and challenges,'' \emph{International Transactions on Electrical Energy Systems}, vol. 2022, no.~1, p. 3358795, 2022.

\bibitem{rana2023applications}
M.~M. Rana, M.~Uddin, M.~R. Sarkar, S.~T. Meraj, G.~Shafiullah, S.~Muyeen, M.~A. Islam, and T.~Jamal, ``Applications of energy storage systems in power grids with and without renewable energy integration—a comprehensive review,'' \emph{Journal of energy storage}, vol.~68, p. 107811, 2023.

\bibitem{rathor2020energy}
S.~K. Rathor and D.~Saxena, ``Energy management system for smart grid: An overview and key issues,'' \emph{International Journal of Energy Research}, vol.~44, no.~6, pp. 4067--4109, 2020.

\bibitem{meliani2021energy}
M.~Meliani, A.~E. Barkany, I.~E. Abbassi, A.~M. Darcherif, and M.~Mahmoudi, ``Energy management in the smart grid: State-of-the-art and future trends,'' \emph{International Journal of Engineering Business Management}, vol.~13, p. 18479790211032920, 2021.

\bibitem{black2022have}
B.~C. Black, \emph{To Have and Have Not: Energy in World History}.\hskip 1em plus 0.5em minus 0.4em\relax Rowman \& Littlefield, 2022.

\bibitem{dickinson2022short}
H.~W. Dickinson, \emph{A short history of the steam engine}.\hskip 1em plus 0.5em minus 0.4em\relax Routledge, 2022.

\bibitem{wellum2023energizing}
C.~Wellum, \emph{Energizing Neoliberalism: The 1970s Energy Crisis and the Making of Modern America}.\hskip 1em plus 0.5em minus 0.4em\relax JHU Press, 2023.

\bibitem{turnbull2022no}
T.~Turnbull, ``‘no solution to the immediate crisis’: The uncertain political economy of energy conservation in 1970s britain,'' \emph{Contemporary European History}, vol.~31, no.~4, pp. 570--592, 2022.

\bibitem{o2021corporate}
E.~O'Shaughnessy, J.~Heeter, C.~Shah, and S.~Koebrich, ``Corporate acceleration of the renewable energy transition and implications for electric grids,'' \emph{Renewable and Sustainable Energy Reviews}, vol. 146, p. 111160, 2021.

\bibitem{o2019smart}
E.~O’Dwyer, I.~Pan, S.~Acha, and N.~Shah, ``Smart energy systems for sustainable smart cities: Current developments, trends and future directions,'' \emph{Applied energy}, vol. 237, pp. 581--597, 2019.

\bibitem{bibri2021novel}
S.~E. Bibri, ``A novel model for data-driven smart sustainable cities of the future: the institutional transformations required for balancing and advancing the three goals of sustainability,'' \emph{Energy Informatics}, vol.~4, no.~1, p.~4, 2021.

\bibitem{zheng2024systematic}
Z.~Zheng, M.~Shafique, X.~Luo, and S.~Wang, ``A systematic review towards integrative energy management of smart grids and urban energy systems,'' \emph{Renewable and Sustainable Energy Reviews}, vol. 189, p. 114023, 2024.

\bibitem{olatunde2024impact}
T.~M. Olatunde, A.~C. Okwandu, D.~O. Akande, and Z.~Q. Sikhakhane, ``The impact of smart grids on energy efficiency: a comprehensive review,'' \emph{Engineering Science \& Technology Journal}, vol.~5, no.~4, pp. 1257--1269, 2024.

\bibitem{bakare2023comprehensive}
M.~S. Bakare, A.~Abdulkarim, M.~Zeeshan, and A.~N. Shuaibu, ``A comprehensive overview on demand side energy management towards smart grids: challenges, solutions, and future direction,'' \emph{Energy Informatics}, vol.~6, no.~1, p.~4, 2023.

\bibitem{poompavai2019control}
T.~Poompavai and M.~Kowsalya, ``Control and energy management strategies applied for solar photovoltaic and wind energy fed water pumping system: A review,'' \emph{Renewable and sustainable energy reviews}, vol. 107, pp. 108--122, 2019.

\bibitem{azuatalam2019energy}
D.~Azuatalam, K.~Paridari, Y.~Ma, M.~F{\"o}rstl, A.~C. Chapman, and G.~Verbi{\v{c}}, ``Energy management of small-scale pv-battery systems: A systematic review considering practical implementation, computational requirements, quality of input data and battery degradation,'' \emph{Renewable and Sustainable Energy Reviews}, vol. 112, pp. 555--570, 2019.

\bibitem{yousefi2019comparison}
M.~Yousefi, A.~Hajizadeh, and M.~N. Soltani, ``A comparison study on stochastic modeling methods for home energy management systems,'' \emph{IEEE Transactions on Industrial Informatics}, vol.~15, no.~8, pp. 4799--4808, 2019.

\bibitem{zou2019survey}
H.~Zou, S.~Mao, Y.~Wang, F.~Zhang, X.~Chen, and L.~Cheng, ``A survey of energy management in interconnected multi-microgrids,'' \emph{IEEE Access}, vol.~7, pp. 72\,158--72\,169, 2019.

\bibitem{cheng2018centralize}
Z.~Cheng, J.~Duan, and M.-Y. Chow, ``To centralize or to distribute: That is the question: A comparison of advanced microgrid management systems,'' \emph{IEEE Industrial Electronics Magazine}, vol.~12, no.~1, pp. 6--24, 2018.

\bibitem{shareef2018review}
H.~Shareef, M.~S. Ahmed, A.~Mohamed, and E.~Al~Hassan, ``Review on home energy management system considering demand responses, smart technologies, and intelligent controllers,'' \emph{Ieee Access}, vol.~6, pp. 24\,498--24\,509, 2018.

\bibitem{hannan2018review}
M.~A. Hannan, M.~Faisal, P.~J. Ker, L.~H. Mun, K.~Parvin, T.~M.~I. Mahlia, and F.~Blaabjerg, ``A review of internet of energy based building energy management systems: Issues and recommendations,'' \emph{IEEE access}, vol.~6, pp. 38\,997--39\,014, 2018.

\bibitem{zia2018microgrids}
M.~F. Zia, E.~Elbouchikhi, and M.~Benbouzid, ``Microgrids energy management systems: A critical review on methods, solutions, and prospects,'' \emph{Applied energy}, vol. 222, pp. 1033--1055, 2018.

\bibitem{vivas2018review}
F.~Vivas, A.~De~las Heras, F.~Segura, and J.~And{\'u}jar, ``A review of energy management strategies for renewable hybrid energy systems with hydrogen backup,'' \emph{Renewable and Sustainable Energy Reviews}, vol.~82, pp. 126--155, 2018.

\bibitem{zakaria2020uncertainty}
A.~Zakaria, F.~B. Ismail, M.~H. Lipu, and M.~A. Hannan, ``Uncertainty models for stochastic optimization in renewable energy applications,'' \emph{Renewable Energy}, vol. 145, pp. 1543--1571, 2020.

\bibitem{khan2019optimal}
M.~W. Khan, J.~Wang, M.~Ma, L.~Xiong, P.~Li, and F.~Wu, ``Optimal energy management and control aspects of distributed microgrid using multi-agent systems,'' \emph{Sustainable Cities and Society}, vol.~44, pp. 855--870, 2019.

\bibitem{salimi2019critical}
S.~Salimi and A.~Hammad, ``Critical review and research roadmap of office building energy management based on occupancy monitoring,'' \emph{Energy and Buildings}, vol. 182, pp. 214--241, 2019.

\bibitem{vlachokostas2020closing}
C.~Vlachokostas, ``Closing the loop between energy production and waste management: A conceptual approach towards sustainable development,'' \emph{Sustainability}, vol.~12, no.~15, p. 5995, 2020.

\bibitem{zhou2016understanding}
K.~Zhou and S.~Yang, ``Understanding household energy consumption behavior: The contribution of energy big data analytics,'' \emph{Renewable and Sustainable Energy Reviews}, vol.~56, pp. 810--819, 2016.

\bibitem{simpeh2022improving}
E.~K. Simpeh, J.-P.~G. Pillay, R.~Ndihokubwayo, and D.~J. Nalumu, ``Improving energy efficiency of hvac systems in buildings: A review of best practices,'' \emph{International Journal of Building Pathology and Adaptation}, vol.~40, no.~2, pp. 165--182, 2022.

\bibitem{gonzalez2018multi}
A.~Gonz{\'a}lez-Briones, F.~De~La~Prieta, M.~S. Mohamad, S.~Omatu, and J.~M. Corchado, ``Multi-agent systems applications in energy optimization problems: A state-of-the-art review,'' \emph{Energies}, vol.~11, no.~8, p. 1928, 2018.

\bibitem{maloberti2022short}
F.~Maloberti and A.~C. Davies, \emph{A short history of circuits and systems}.\hskip 1em plus 0.5em minus 0.4em\relax CRC Press, 2022.

\bibitem{katyara2019monitoring}
S.~Katyara, M.~A. Shah, B.~S. Chowdhary, F.~Akhtar, and G.~A. Lashari, ``Monitoring, control and energy management of smart grid system via wsn technology through scada applications,'' \emph{Wireless Personal Communications}, vol. 106, no.~4, pp. 1951--1968, 2019.

\bibitem{husin2021critical}
H.~Husin, M.~Zaki \emph{et~al.}, ``A critical review of the integration of renewable energy sources with various technologies,'' \emph{Protection and control of modern power systems}, vol.~6, no.~1, pp. 1--18, 2021.

\bibitem{al2022sustainable}
A.~Q. Al-Shetwi, ``Sustainable development of renewable energy integrated power sector: Trends, environmental impacts, and recent challenges,'' \emph{Science of The Total Environment}, vol. 822, p. 153645, 2022.

\bibitem{hossain2021metrics}
E.~Hossain, S.~Roy, N.~Mohammad, N.~Nawar, and D.~R. Dipta, ``Metrics and enhancement strategies for grid resilience and reliability during natural disasters,'' \emph{Applied energy}, vol. 290, p. 116709, 2021.

\bibitem{xu2021resilience}
L.~Xu, Q.~Guo, Y.~Sheng, S.~Muyeen, and H.~Sun, ``On the resilience of modern power systems: A comprehensive review from the cyber-physical perspective,'' \emph{Renewable and Sustainable Energy Reviews}, vol. 152, p. 111642, 2021.

\bibitem{gold2020leveraging}
R.~Gold, C.~Waters, and D.~York, ``Leveraging advanced metering infrastructure to save energy,'' \emph{American Council for an Energy-Efficient Economy (ACEEE), Washington DC}, 2020.

\bibitem{marimuthu2018development}
K.~P. Marimuthu, D.~Durairaj, and S.~Karthik~Srinivasan, ``Development and implementation of advanced metering infrastructure for efficient energy utilization in smart grid environment,'' \emph{International Transactions on Electrical Energy Systems}, vol.~28, no.~3, p. e2504, 2018.

\bibitem{khan2019secure}
S.~Khan, R.~Khan, and A.~H. Al-Bayatti, ``Secure communication architecture for dynamic energy management in smart grid,'' \emph{IEEE Power and Energy Technology Systems Journal}, vol.~6, no.~1, pp. 47--58, 2019.

\bibitem{dileep2020survey}
G.~Dileep, ``A survey on smart grid technologies and applications,'' \emph{Renewable energy}, vol. 146, pp. 2589--2625, 2020.

\bibitem{chaves2022development}
T.~R. Chaves, M.~A. Martins, K.~A. Martins, and A.~F. Macedo, ``Development of an automated distribution grid with the application of new technologies,'' \emph{IEEE Access}, vol.~10, pp. 9431--9445, 2022.

\bibitem{ge2022smart}
L.~Ge, Y.~Li, Y.~Li, J.~Yan, and Y.~Sun, ``Smart distribution network situation awareness for high-quality operation and maintenance: a brief review,'' \emph{Energies}, vol.~15, no.~3, p. 828, 2022.

\bibitem{marques2024artificial}
P.~C. Marques and P.~A. Oliveira, ``Artificial intelligence technologies applied to smart grids and management,'' 2024.

\bibitem{saleem2023integrating}
M.~U. Saleem, M.~Shakir, M.~R. Usman, M.~H.~T. Bajwa, N.~Shabbir, P.~Shams~Ghahfarokhi, and K.~Daniel, ``Integrating smart energy management system with internet of things and cloud computing for efficient demand side management in smart grids,'' \emph{Energies}, vol.~16, no.~12, p. 4835, 2023.

\bibitem{hussain2021multi}
S.~Hussain, C.~Z. El-Bayeh, C.~Lai, and U.~Eicker, ``Multi-level energy management systems toward a smarter grid: A review,'' \emph{IEEE Access}, vol.~9, pp. 71\,994--72\,016, 2021.

\bibitem{datchanamoorthy2011optimal}
S.~Datchanamoorthy, S.~Kumar, Y.~Ozturk, and G.~Lee, ``Optimal time-of-use pricing for residential load control,'' in \emph{2011 IEEE International Conference on Smart Grid Communications (SmartGridComm)}.\hskip 1em plus 0.5em minus 0.4em\relax IEEE, 2011, pp. 375--380.

\bibitem{el2020novel}
C.~Z. El-Bayeh, U.~Eicker, K.~Alzaareer, B.~Brahmi, and M.~Zellagui, ``A novel data-energy management algorithm for smart transformers to optimize the total load demand in smart homes,'' \emph{Energies}, vol.~13, no.~18, p. 4984, 2020.

\bibitem{herter2007residential}
K.~Herter, ``Residential implementation of critical-peak pricing of electricity,'' \emph{Energy policy}, vol.~35, no.~4, pp. 2121--2130, 2007.

\bibitem{borenstein2008equity}
S.~Borenstein, ``Equity effects of increasing-block electricity pricing,'' 2008.

\bibitem{ericson2009direct}
T.~Ericson, ``Direct load control of residential water heaters,'' \emph{Energy Policy}, vol.~37, no.~9, pp. 3502--3512, 2009.

\bibitem{hashmi2021internet}
S.~A. Hashmi, C.~F. Ali, and S.~Zafar, ``Internet of things and cloud computing-based energy management system for demand side management in smart grid,'' \emph{International Journal of Energy Research}, vol.~45, no.~1, pp. 1007--1022, 2021.

\bibitem{zhao2020smart}
S.~Zhao, F.~Li, H.~Li, R.~Lu, S.~Ren, H.~Bao, J.-H. Lin, and S.~Han, ``Smart and practical privacy-preserving data aggregation for fog-based smart grids,'' \emph{IEEE Transactions on Information Forensics and Security}, vol.~16, pp. 521--536, 2020.

\bibitem{umapathy2023machine}
K.~Umapathy, T.~Dinesh~Kumar, G.~Poojitha, D.~Khyathi~Sri, C.~Pavaneeswar, and C.~Amannah, ``Machine learning applications for the smart grid,'' in \emph{Data Analytics for Smart Grids Applications—A Key to Smart City Development}.\hskip 1em plus 0.5em minus 0.4em\relax Springer, 2023, pp. 251--270.

\bibitem{yang2019bayesian}
Y.~Yang, W.~Li, T.~A. Gulliver, and S.~Li, ``Bayesian deep learning-based probabilistic load forecasting in smart grids,'' \emph{IEEE Transactions on Industrial Informatics}, vol.~16, no.~7, pp. 4703--4713, 2019.

\bibitem{berghout2022machine}
T.~Berghout, M.~Benbouzid, and S.~Muyeen, ``Machine learning for cybersecurity in smart grids: A comprehensive review-based study on methods, solutions, and prospects,'' \emph{International Journal of Critical Infrastructure Protection}, vol.~38, p. 100547, 2022.

\bibitem{siniosoglou2021unified}
I.~Siniosoglou, P.~Radoglou-Grammatikis, G.~Efstathopoulos, P.~Fouliras, and P.~Sarigiannidis, ``A unified deep learning anomaly detection and classification approach for smart grid environments,'' \emph{IEEE Transactions on Network and Service Management}, vol.~18, no.~2, pp. 1137--1151, 2021.

\bibitem{kataray2023integration}
T.~Kataray, B.~Nitesh, B.~Yarram, S.~Sinha, E.~Cuce, S.~Shaik, P.~Vigneshwaran, and A.~Roy, ``Integration of smart grid with renewable energy sources: Opportunities and challenges--a comprehensive review,'' \emph{Sustainable Energy Technologies and Assessments}, vol.~58, p. 103363, 2023.

\bibitem{vijayapriya2011smart}
T.~Vijayapriya and D.~P. Kothari, ``Smart grid: an overview,'' \emph{Smart Grid and Renewable Energy}, vol.~2, no.~4, pp. 305--311, 2011.

\bibitem{chong2020review}
A.~T.~Y. Chong, M.~A. Mahmoud, F.-C. Lim, and H.~Kasim, ``A review of smart grid technology, components, and implementation,'' in \emph{2020 8th International Conference on Information Technology and Multimedia (ICIMU)}.\hskip 1em plus 0.5em minus 0.4em\relax IEEE, 2020, pp. 166--169.

\bibitem{kabeyi2023smart}
M.~J.~B. Kabeyi and O.~A. Olanrewaju, ``Smart grid technologies and application in the sustainable energy transition: a review,'' \emph{International Journal of Sustainable Energy}, vol.~42, no.~1, pp. 685--758, 2023.

\bibitem{solomon2022outage}
E.~Solomon, I.~G. Hagos, and B.~Khan, ``Outage management in a smart distribution grid integrated with pev and its v2b applications,'' in \emph{Active Electrical Distribution Network}.\hskip 1em plus 0.5em minus 0.4em\relax Elsevier, 2022, pp. 217--226.

\bibitem{ukoba2023geographic}
M.~Ukoba, E.~Diemuodeke, T.~Briggs, M.~Imran, K.~Owebor, and C.~Nwachukwu, ``Geographic information systems (gis) approach for assessing the biomass energy potential and identification of appropriate biomass conversion technologies in nigeria,'' \emph{Biomass and Bioenergy}, vol. 170, p. 106726, 2023.

\bibitem{phing2024brief}
C.~C. Phing, T.~S. Kiong, S.~P. Koh, T.~Abedin, Y.~C. Tak, T.~Yusaf, M.~W. Yaw \emph{et~al.}, ``A brief review on ancillary services from advanced metering infrastructure (asami) for distributed renewable energy network,'' \emph{Journal of Advanced Research in Applied Sciences and Engineering Technology}, vol.~41, no.~2, pp. 43--61, 2024.

\bibitem{chen2023control}
Z.~Chen, A.~M. Amani, X.~Yu, and M.~Jalili, ``Control and optimisation of power grids using smart meter data: A review,'' \emph{Sensors}, vol.~23, no.~4, p. 2118, 2023.

\bibitem{alomar2023iot}
M.~A. Alomar, ``An iot based smart grid system for advanced cooperative transmission and communication,'' \emph{Physical Communication}, vol.~58, p. 102069, 2023.

\bibitem{sajid2024multi}
A.~H. Sajid, A.~Altamimi, S.~A.~A. Kazmi, and Z.~A. Khan, ``Multi-micro grid system reinforcement across deregulated markets, energy resources scheduling and demand side management using a multi-agent-based optimization in smart grid paradigm,'' \emph{IEEE Access}, 2024.

\bibitem{yarahmadi2024improving}
H.~Yarahmadi, H.~Navidi, and M.~Challenger, ``Improving the resource allocation in iot systems based on the integration of reinforcement learning and rule-based approaches in multi-agent systems,'' in \emph{2024 8th International Conference on Smart Cities, Internet of Things and Applications (SCIoT)}.\hskip 1em plus 0.5em minus 0.4em\relax IEEE, 2024, pp. 135--141.

\bibitem{panda2023recent}
A.~Panda, A.~K. Dauda, H.~Chua, R.~R. Tan, and K.~B. Aviso, ``Recent advances in the integration of renewable energy sources and storage facilities with hybrid power systems,'' \emph{Cleaner Engineering and Technology}, vol.~12, p. 100598, 2023.

\bibitem{etukudoh2024electrical}
E.~A. Etukudoh, A.~Fabuyide, K.~I. Ibekwe, S.~Sonko, and V.~I. Ilojianya, ``Electrical engineering in renewable energy systems: a review of design and integration challenges,'' \emph{Engineering Science \& Technology Journal}, vol.~5, no.~1, pp. 231--244, 2024.

\bibitem{barman2023renewable}
P.~Barman, L.~Dutta, S.~Bordoloi, A.~Kalita, P.~Buragohain, S.~Bharali, and B.~Azzopardi, ``Renewable energy integration with electric vehicle technology: A review of the existing smart charging approaches,'' \emph{Renewable and Sustainable Energy Reviews}, vol. 183, p. 113518, 2023.

\bibitem{yadav2024smart}
P.~K. Yadav, M.~Biswal, and H.~Vemuganti, ``Smart meter data management challenges,'' in \emph{Smart Metering}.\hskip 1em plus 0.5em minus 0.4em\relax Elsevier, 2024, pp. 221--256.

\bibitem{shaukat2023decentralized}
N.~Shaukat, M.~R. Islam, M.~M. Rahman, B.~Khan, B.~Ullah, S.~M. Ali, and A.~Fekih, ``Decentralized, democratized, and decarbonized future electric power distribution grids: a survey on the paradigm shift from the conventional power system to micro grid structures,'' \emph{IEEE Access}, vol.~11, pp. 60\,957--60\,987, 2023.

\bibitem{moreno2021comprehensive}
J.~J. Moreno~Escobar, O.~Morales~Matamoros, R.~Tejeida~Padilla, I.~Lina~Reyes, and H.~Quintana~Espinosa, ``A comprehensive review on smart grids: Challenges and opportunities,'' \emph{Sensors}, vol.~21, no.~21, p. 6978, 2021.

\bibitem{goudarzi2022survey}
A.~Goudarzi, F.~Ghayoor, M.~Waseem, S.~Fahad, and I.~Traore, ``A survey on iot-enabled smart grids: emerging, applications, challenges, and outlook,'' \emph{Energies}, vol.~15, no.~19, p. 6984, 2022.

\bibitem{osti_1804705}
\BIBentryALTinterwordspacing
P.~Denholm, D.~J. Arent, S.~F. Baldwin, D.~E. Bilello, G.~L. Brinkman, J.~M. Cochran, W.~J. Cole, B.~Frew, V.~Gevorgian, J.~Heeter, B.-M.~S. Hodge, B.~Kroposki, T.~Mai, M.~J. O'Malley, B.~Palmintier, D.~Steinberg, and Y.~Zhang, ``The challenges of achieving a 100
\BIBentrySTDinterwordspacing

\bibitem{pandiyan2023technological}
P.~Pandiyan, S.~Saravanan, K.~Usha, R.~Kannadasan, M.~H. Alsharif, and M.-K. Kim, ``Technological advancements toward smart energy management in smart cities,'' \emph{Energy Reports}, vol.~10, pp. 648--677, 2023.

\bibitem{ezeigweneme2024smart}
C.~A. Ezeigweneme, C.~N. Nwasike, A.~Adefemi, A.~O. Adegbite, and J.~O. Gidiagba, ``Smart grids in industrial paradigms: a review of progress, benefits, and maintenance implications: analyzing the role of smart grids in predictive maintenance and the integration of renewable energy sources, along with their overall impact on the industri,'' \emph{Engineering Science \& Technology Journal}, vol.~5, no.~1, pp. 1--20, 2024.

\bibitem{maruf2020adaptation}
M.~H. Maruf, M.~A. ul~Haq, S.~K. Dey, A.~Al~Mansur, and A.~Shihavuddin, ``Adaptation for sustainable implementation of smart grid in developing countries like bangladesh,'' \emph{Energy Reports}, vol.~6, pp. 2520--2530, 2020.

\bibitem{khalid2024smart}
M.~Khalid, ``Smart grids and renewable energy systems: Perspectives and grid integration challenges,'' \emph{Energy Strategy Reviews}, vol.~51, p. 101299, 2024.

\bibitem{bou2013communication}
E.~Bou-Harb, C.~Fachkha, M.~Pourzandi, M.~Debbabi, and C.~Assi, ``Communication security for smart grid distribution networks,'' \emph{IEEE Communications Magazine}, vol.~51, no.~1, pp. 42--49, 2013.

\bibitem{he2016cyber}
H.~He and J.~Yan, ``Cyber-physical attacks and defences in the smart grid: a survey,'' \emph{IET Cyber-Physical Systems: Theory \& Applications}, vol.~1, no.~1, pp. 13--27, 2016.

\bibitem{leiva2016smart}
J.~Leiva, A.~Palacios, and J.~A. Aguado, ``Smart metering trends, implications and necessities: A policy review,'' \emph{Renewable and sustainable energy reviews}, vol.~55, pp. 227--233, 2016.

\bibitem{anzalchi2015survey}
A.~Anzalchi and A.~Sarwat, ``A survey on security assessment of metering infrastructure in smart grid systems,'' in \emph{SoutheastCon 2015}.\hskip 1em plus 0.5em minus 0.4em\relax IEEE, 2015, pp. 1--4.

\bibitem{wei2018review}
L.~Wei, L.~P. Rondon, A.~Moghadasi, and A.~I. Sarwat, ``Review of cyber-physical attacks and counter defense mechanisms for advanced metering infrastructure in smart grid,'' in \emph{2018 IEEE/PES Transmission and Distribution Conference and Exposition (T\&D)}.\hskip 1em plus 0.5em minus 0.4em\relax IEEE, 2018, pp. 1--9.

\bibitem{wang2011analysis}
Y.~Wang, D.~Ruan, D.~Gu, J.~Gao, D.~Liu, J.~Xu, F.~Chen, F.~Dai, and J.~Yang, ``Analysis of smart grid security standards,'' in \emph{2011 IEEE International Conference on Computer Science and Automation Engineering}, vol.~4.\hskip 1em plus 0.5em minus 0.4em\relax IEEE, 2011, pp. 697--701.

\bibitem{leszczyna2018cybersecurity}
R.~Leszczyna, ``Cybersecurity and privacy in standards for smart grids--a comprehensive survey,'' \emph{Computer Standards \& Interfaces}, vol.~56, pp. 62--73, 2018.

\bibitem{gunduz2020cyber}
M.~Z. Gunduz and R.~Das, ``Cyber-security on smart grid: Threats and potential solutions,'' \emph{Computer networks}, vol. 169, p. 107094, 2020.

\bibitem{srivastava2021emerging}
A.~Srivastava and A.~Agarwal, ``Emerging technology iot and ot: overview, security threats, attacks and countermeasures,'' \emph{IJERT}, vol.~10, no.~7, pp. 86--93, 2021.

\bibitem{rajendran2021comprehensive}
G.~Rajendran, C.~A. Vaithilingam, N.~Misron, K.~Naidu, and M.~R. Ahmed, ``A comprehensive review on system architecture and international standards for electric vehicle charging stations,'' \emph{Journal of Energy Storage}, vol.~42, p. 103099, 2021.

\bibitem{liberati2021review}
F.~Liberati, E.~Garone, and A.~Di~Giorgio, ``Review of cyber-physical attacks in smart grids: a system-theoretic perspective,'' \emph{Electronics}, vol.~10, no.~10, p. 1153, 2021.

\bibitem{hasan2023review}
M.~K. Hasan, A.~A. Habib, Z.~Shukur, F.~Ibrahim, S.~Islam, and M.~A. Razzaque, ``Review on cyber-physical and cyber-security system in smart grid: Standards, protocols, constraints, and recommendations,'' \emph{Journal of network and computer applications}, vol. 209, p. 103540, 2023.

\bibitem{kim2023smart}
Y.~Kim, S.~Hakak, and A.~Ghorbani, ``Smart grid security: Attacks and defence techniques,'' \emph{IET Smart Grid}, vol.~6, no.~2, pp. 103--123, 2023.

\bibitem{7454291}
S.~Shapsough, F.~Qatan, R.~Aburukba, F.~Aloul, and A.~R. Al~Ali, ``Smart grid cyber security: Challenges and solutions,'' in \emph{2015 International Conference on Smart Grid and Clean Energy Technologies (ICSGCE)}, 2015, pp. 170--175.

\bibitem{mohammadi2021emerging}
F.~Mohammadi, ``Emerging challenges in smart grid cybersecurity enhancement: A review,'' \emph{Energies}, vol.~14, no.~5, p. 1380, 2021.

\bibitem{faquir2021cybersecurity}
D.~Faquir, N.~Chouliaras, V.~Sofia, K.~Olga, and L.~Maglaras, ``Cybersecurity in smart grids, challenges and solutions,'' \emph{AIMS Electronics and Electrical Engineering}, vol.~5, no.~1, pp. 24--37, 2021.

\bibitem{ortega2023review}
I.~Ortega-Fernandez and F.~Liberati, ``A review of denial of service attack and mitigation in the smart grid using reinforcement learning,'' \emph{Energies}, vol.~16, no.~2, p. 635, 2023.

\bibitem{akhtar2018malware}
T.~Akhtar, B.~B. Gupta, and S.~Yamaguchi, ``Malware propagation effects on scada system and smart power grid,'' in \emph{2018 IEEE International Conference on Consumer Electronics (ICCE)}.\hskip 1em plus 0.5em minus 0.4em\relax IEEE, 2018, pp. 1--6.

\bibitem{ding2022cyber}
J.~Ding, A.~Qammar, Z.~Zhang, A.~Karim, and H.~Ning, ``Cyber threats to smart grids: Review, taxonomy, potential solutions, and future directions,'' \emph{Energies}, vol.~15, no.~18, p. 6799, 2022.

\bibitem{patil2020grid}
H.~Patil and V.~N. Kalkhambkar, ``Grid integration of electric vehicles for economic benefits: A review,'' \emph{Journal of Modern Power Systems and Clean Energy}, vol.~9, no.~1, pp. 13--26, 2020.

\bibitem{wu2021benefits}
W.~Wu and B.~Lin, ``Benefits of electric vehicles integrating into power grid,'' \emph{Energy}, vol. 224, p. 120108, 2021.

\bibitem{venegas2021active}
F.~G. Venegas, M.~Petit, and Y.~Perez, ``Active integration of electric vehicles into distribution grids: Barriers and frameworks for flexibility services,'' \emph{Renewable and Sustainable Energy Reviews}, vol. 145, p. 111060, 2021.

\bibitem{syed2020smart}
D.~Syed, A.~Zainab, A.~Ghrayeb, S.~S. Refaat, H.~Abu-Rub, and O.~Bouhali, ``Smart grid big data analytics: Survey of technologies, techniques, and applications,'' \emph{IEEE Access}, vol.~9, pp. 59\,564--59\,585, 2020.

\bibitem{omitaomu2021artificial}
O.~A. Omitaomu and H.~Niu, ``Artificial intelligence techniques in smart grid: A survey,'' \emph{Smart Cities}, vol.~4, no.~2, pp. 548--568, 2021.

\bibitem{shi2020artificial}
Z.~Shi, W.~Yao, Z.~Li, L.~Zeng, Y.~Zhao, R.~Zhang, Y.~Tang, and J.~Wen, ``Artificial intelligence techniques for stability analysis and control in smart grids: Methodologies, applications, challenges and future directions,'' \emph{Applied Energy}, vol. 278, p. 115733, 2020.

\bibitem{ali2020state}
S.~S. Ali and B.~J. Choi, ``State-of-the-art artificial intelligence techniques for distributed smart grids: A review,'' \emph{Electronics}, vol.~9, no.~6, p. 1030, 2020.

\bibitem{9395437}
M.~Massaoudi, H.~Abu-Rub, S.~S. Refaat, I.~Chihi, and F.~S. Oueslati, ``Deep learning in smart grid technology: A review of recent advancements and future prospects,'' \emph{IEEE Access}, vol.~9, pp. 54\,558--54\,578, 2021.

\bibitem{9293294}
Z.~Wang, M.~Ogbodo, H.~Huang, C.~Qiu, M.~Hisada, and A.~B. Abdallah, ``Aebis: Ai-enabled blockchain-based electric vehicle integration system for power management in smart grid platform,'' \emph{IEEE Access}, vol.~8, pp. 226\,409--226\,421, 2020.

\bibitem{9084590}
S.~Azad, F.~Sabrina, and S.~Wasimi, ``Transformation of smart grid using machine learning,'' in \emph{2019 29th Australasian Universities Power Engineering Conference (AUPEC)}, 2019, pp. 1--6.

\bibitem{zhao2023blockchain}
W.~Zhao, Q.~Qi, J.~Zhou, and X.~Luo, ``Blockchain-based applications for smart grids: An umbrella review,'' \emph{Energies}, vol.~16, no.~17, p. 6147, 2023.

\bibitem{pallonetto2019demand}
F.~Pallonetto, M.~De~Rosa, F.~Milano, and D.~P. Finn, ``Demand response algorithms for smart-grid ready residential buildings using machine learning models,'' \emph{Applied energy}, vol. 239, pp. 1265--1282, 2019.

\bibitem{9765492}
T.~Alquthami, M.~Zulfiqar, M.~Kamran, A.~H. Milyani, and M.~B. Rasheed, ``A performance comparison of machine learning algorithms for load forecasting in smart grid,'' \emph{IEEE Access}, vol.~10, pp. 48\,419--48\,433, 2022.

\bibitem{el2014smart}
M.~E. El-Hawary, ``The smart grid—state-of-the-art and future trends,'' \emph{Electric Power Components and Systems}, vol.~42, no. 3-4, pp. 239--250, 2014.

\bibitem{atitallah2020leveraging}
S.~B. Atitallah \emph{et~al.}, ``Leveraging deep learning and iot big data analytics to support the smart cities development: Review and future directions,'' \emph{Computer Science Review}, vol.~38, p. 100303, 2020.

\bibitem{kotsiopoulos2021machine}
T.~Kotsiopoulos \emph{et~al.}, ``Machine learning and deep learning in smart manufacturing: The smart grid paradigm,'' \emph{Computer Science Review}, vol.~40, p. 100341, 2021.

\bibitem{mollah2020blockchain}
M.~B. Mollah \emph{et~al.}, ``Blockchain for future smart grid: A comprehensive survey,'' \emph{IEEE Internet of Things Journal}, vol.~8, no.~1, pp. 18--43, 2020.

\bibitem{ahl2020exploring}
A.~Ahl \emph{et~al.}, ``Exploring blockchain for the energy transition: Opportunities and challenges based on a case study in japan,'' \emph{Renewable and Sustainable Energy Reviews}, vol. 117, p. 109488, 2020.

\bibitem{yapa2021survey}
C.~Yapa, C.~De~Alwis, M.~Liyanage, and J.~Ekanayake, ``Survey on blockchain for future smart grids: Technical aspects, applications, integration challenges and future research,'' \emph{Energy Reports}, vol.~7, pp. 6530--6564, 2021.

\bibitem{smartgrids2008strategic}
{European Technology Platform SmartGrids}, \emph{Strategic Deployment Document for Europe's Electricity Networks of the Future}.\hskip 1em plus 0.5em minus 0.4em\relax Brussels: European Technology Platform SmartGrids, 2008.

\bibitem{butt2021recent}
O.~M. Butt, M.~Zulqarnain, and T.~M. Butt, ``Recent advancement in smart grid technology: Future prospects in the electrical power network,'' \emph{Ain Shams Engineering Journal}, vol.~12, no.~1, pp. 687--695, 2021.

\end{thebibliography}




\begin{IEEEbiography}[{\includegraphics[width=1in,height=1.25in,clip,keepaspectratio]{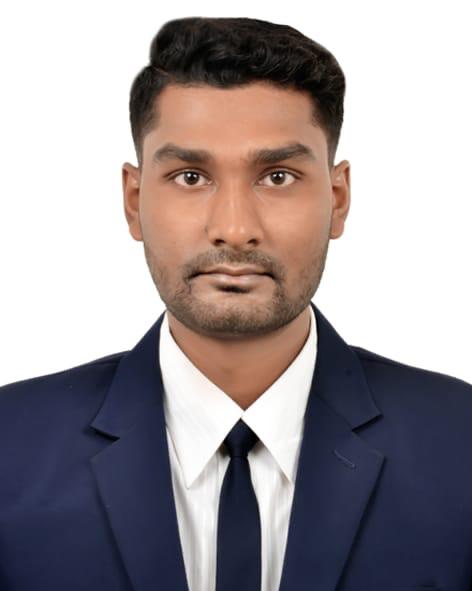}}]{BISWAS PARAG} worked as an Officer (Electrical) in Supply Chain Management at Sajeeb Group from February 2021 to August 2022, where he developed and implemented supply chain strategies. He is currently pursuing a Master of Science in Engineering Management at Westcliff University (August 2022 - Present) and holds a Bachelor of Science in Electrical and Electronic Engineering from Daffodil International University, completed in January 2021, with a GPA of 3.15. Additionally, he conducted undergraduate research from July 2020 to January 2021, focusing on "Biomass Resource Analyses and Future Bioenergy Scenarios" under the supervision of Dr. Arnob Ghosh.
\end{IEEEbiography}

\begin{IEEEbiography}[{\includegraphics[width=1in,height=1.25in,clip,keepaspectratio]{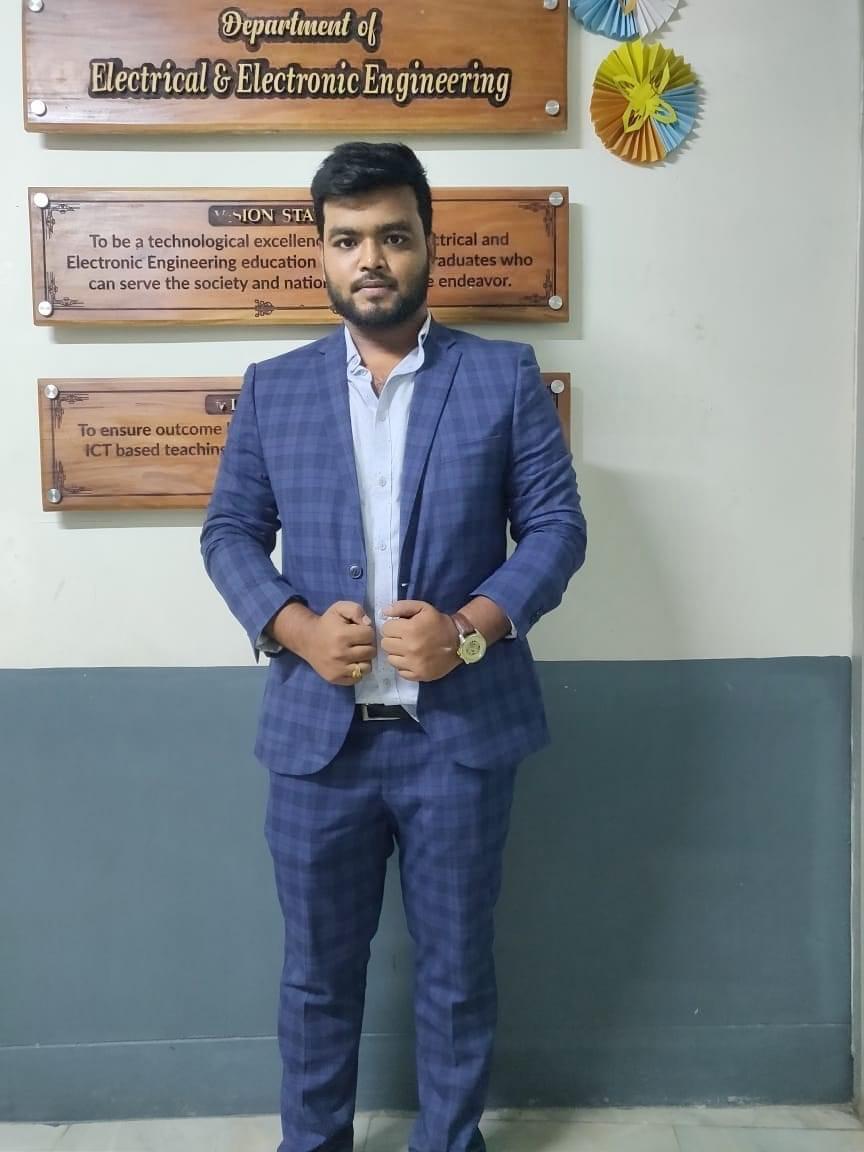}}]{Rashid Abdur} is currently an Electrical Engineer at Gazi Electric Co. since March 2021. He earned his BSc in Electrical and Electronic Engineering from Daffodil International University in January 2021, where he conducted research for his thesis titled "Biomass Resource Analyses and Future Bioenergy Scenarios." Additionally, he completed a Diploma in Engineering from MAWTS Institute of Technology in December 2016.
\end{IEEEbiography}

\begin{IEEEbiography}[{\includegraphics[width=1in,height=1.25in,clip,keepaspectratio]{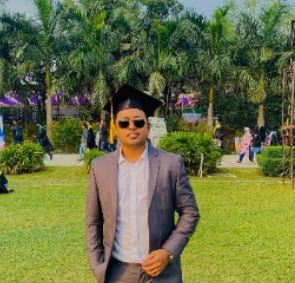}}]{Masum Abdullah Al} has a diverse background in software engineering and education. He completed his Master of Science in Information Technology from Westcliff University in May 2020. He is currently pursuing a Bachelor of Science in Computer Science at Daffodil International University, expected to graduate in November 2024. His professional experience includes working as a Junior Software Engineer at Intelle Hub Inc. from February 2021 to August 2022, an Instructor at Defence Care Academy from April 2020 to January 2021, and an intern in mobile application development at Grameen Solutions Limited from January to March 2020. Additionally, he has engaged in research and system development for the Defence Care Academy, focusing on analyzing and modifying existing software and constructing end-user applications. 
\end{IEEEbiography}

\begin{IEEEbiography}[{\includegraphics[width=1in,height=1.25in,clip,keepaspectratio]{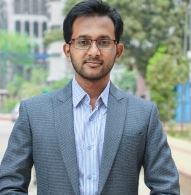}}]{MD Abdullah Al Nasim} received his Bachelor's degree in Computer Science and Engineering from Ahsanullah University of Science and Technology. His research interests include AI, Machine Learning, and Data Science, with a focus on practical applications and innovation. He has been recognized with awards such as the BASIS National ICT Awards, National Hackathon Winner, and an APICTA nomination. He is also the founder of several educational platforms that aim to enhance technical skills and promote STEM education.
\end{IEEEbiography}

\begin{IEEEbiography}[{\includegraphics[width=1in,height=1.25in,clip,keepaspectratio]{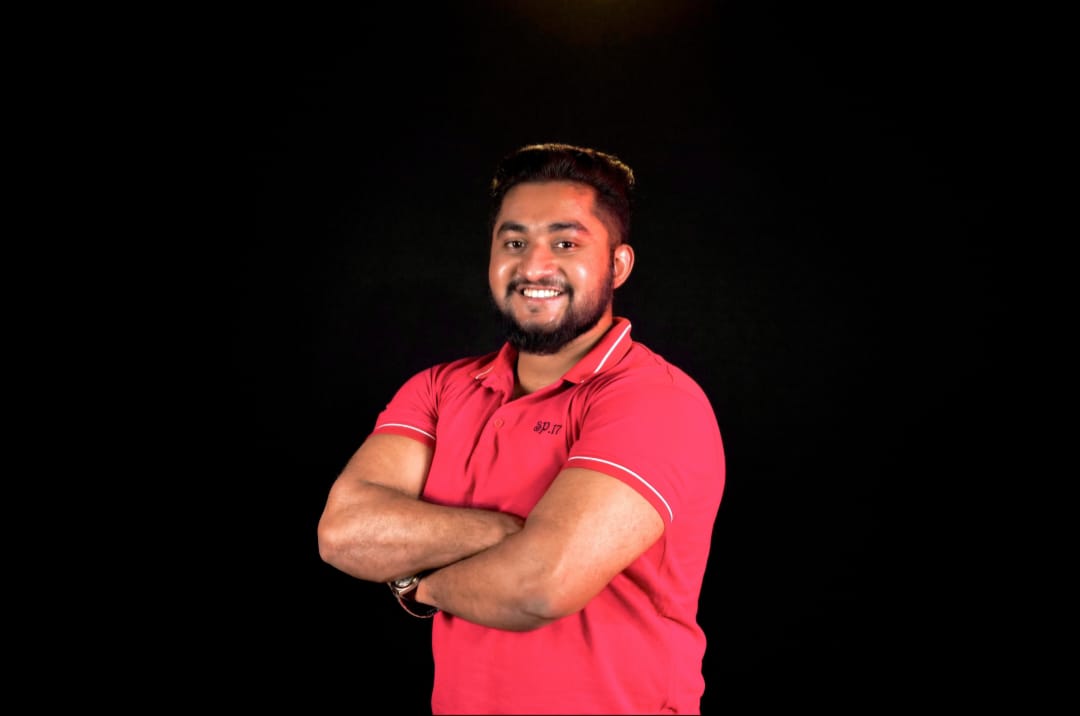}}]{A.S.M ANAS FERDOUS} was born in Dinajpur, Bangladesh, in 2001. He is currently completing the B.Sc. degree in Biomedical Engineering at the Bangladesh University of Engineering and Technology (BUET).

Since 2019, he has been working on AI-based and mHealth applications, having completed two national-level mHealth applications titled DengueDrop and BurnFlow. He has published several research papers and is actively involved in new projects. His research interests include mHealth, telemedicine, artificial intelligence, and biomaterials.

Mr. Ferdous is the Founding Chairman of one of Bangladesh's largest EdTech platforms, Hulkenstein. 
\end{IEEEbiography}

\begin{IEEEbiography}[{\includegraphics[width=1in,height=1.25in,clip,keepaspectratio]{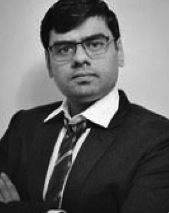}}]{ Gupta Kishor Datta}  The University of Memphis awarded Senior Member of IEEE a Ph.D. He is currently an Assistant Professor at Clark Atlanta University in Atlanta, Georgia's Department of Cyber-Physical Systems. Particularly in the area of adversarial machine learning, Dr. Gupta has made substantial contributions to the field, including one patent and multiple peer-reviewed papers. His areas of interest in study are computer vision, computer security, and bioinspired algorithms. As a member of the program committee for the Flagship Artificial Intelligence Conference AAAI-23, he is also actively involved in the academic community.

\end{IEEEbiography}

\begin{IEEEbiography}[{\includegraphics[width=1in,height=1.25in,clip,keepaspectratio]{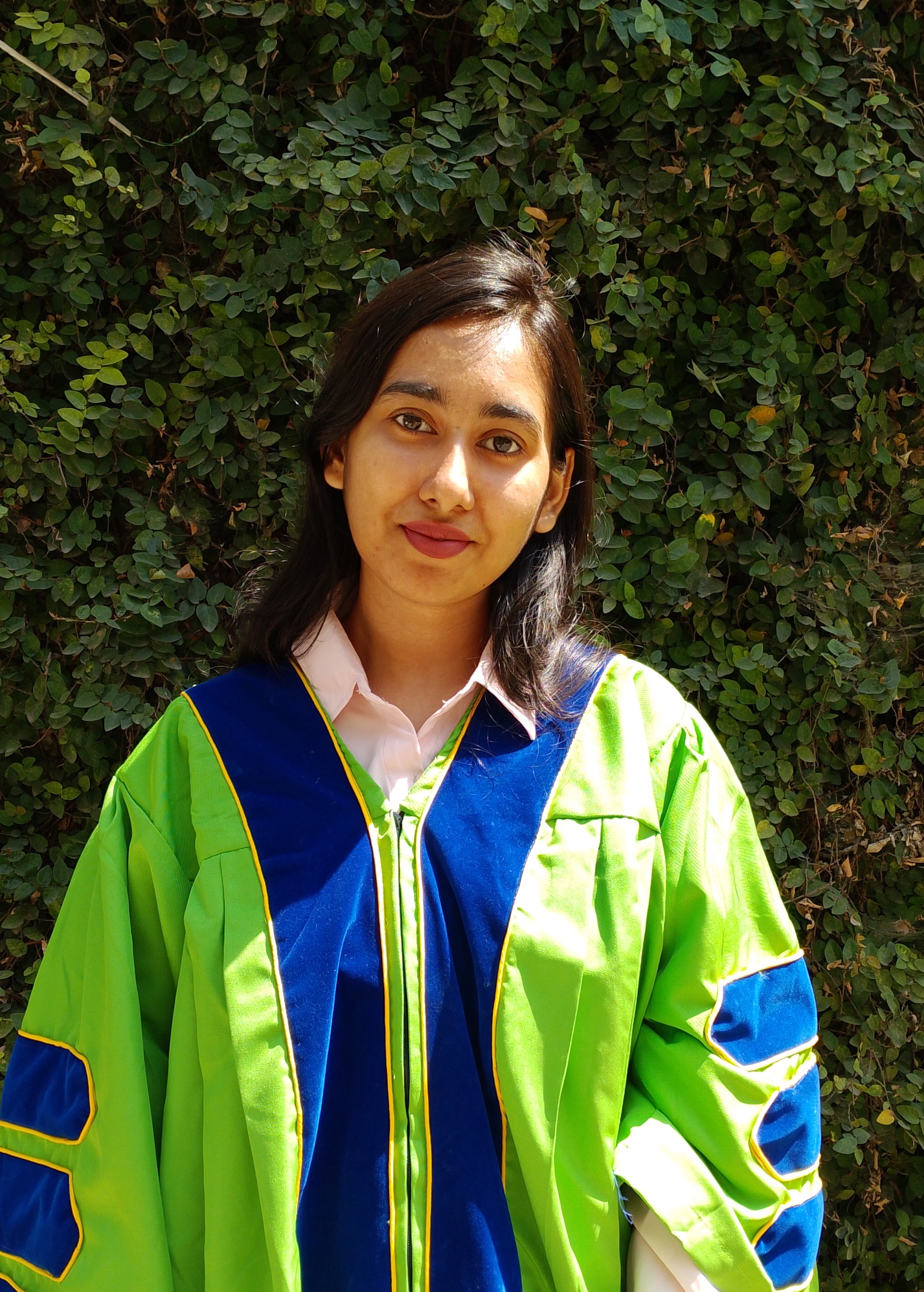}}]{Biswas Angona} received the B.Sc. degree in Electrical and Telecommunication
Engineering from Chittagong University of Engineering and Technology, Bangladesh, in 2022. She is currently looking for opportunity to pursue the Ph.D. 
degree in AI domain.

From 2022 to 2024, she played the role of a Lecture and Instructor for different renowned institutions (Daffodil International University, Presidency University, Pioneer Alpha). Besides this job she also continued her job as a researcher after graduation. 

Ms. Angona's awards and honors include the various IEEE recognition and Dean Award for excellent result in B.Sc. degree. She was member of IEEE since 2019. 
\end{IEEEbiography}

\EOD

\end{document}